\documentclass{article}
\usepackage{arxiv}
\usepackage{amsmath} 
\usepackage[utf8]{inputenc} 
\usepackage[T1]{fontenc}    
\usepackage{hyperref}       
\usepackage{url}            
\usepackage{booktabs}       
\usepackage{amsfonts}       
\usepackage{nicefrac}       
\usepackage{microtype}      
\usepackage{lipsum}		
\usepackage{graphicx}
\usepackage{natbib}
\usepackage{doi}
\usepackage{xcolor}
\usepackage{float}
\usepackage{bm}
\usepackage{algorithm}
\usepackage{algorithmicx}
\usepackage{algpseudocode}
\usepackage{makecell} 
\usepackage{arydshln}
\usepackage{comment}
\usepackage{changepage} 
\usepackage{multirow} 

\title{\fontsize{16pt}{16pt}\selectfont A Physics-Informed Meta-Learning Framework \\ for the Continuous Solution of Parametric PDEs \\ on Arbitrary Geometries}

\author{ 
    \href{https://www.cee.ed.tum.de/en/st/members/reza-najian/}{\includegraphics[scale=0.06]{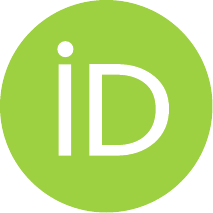}\hspace{1mm}Reza Najian Asl} \\
    	\texttt{reza.najian-asl@tum.de} \\
	\And
	\href{https://muramatsu.mech.keio.ac.jp/students/yusuke-yamazaki/}{\includegraphics[scale=0.06]{orcid.pdf}\hspace{1mm}Yusuke Yamazaki} \\
	Graduate School of Science and Technology, \\ Keio University, Hiyoshi3‑14‑1,\\ Kohoku‑ku, Yokohama 223‑8522, Japan\\
	\texttt{yusuke.yamazaki.0615@keio.jp} \\
	\And
	\href{https://www.linkedin.com/in/kianoosh-taghikhani-8ba5586b/?originalSubdomain=de}{\includegraphics[scale=0.06]{orcid.pdf}\hspace{1mm}Kianoosh Taghikhani} \\
	Access e.V. \\ Intzestr. 5, D-52072, Aachen, Germany\\
	\texttt{kianoosh.taghikhani@rwth-aachen.de} \\
	\And
	\href{https://www.st.keio.ac.jp/en/tprofile/mech/muramatsu.html}{\includegraphics[scale=0.06]{orcid.pdf}\hspace{1mm}Mayu Muramatsu} \\
	Department of Mechanical Engineering,\\ Keio University, Hiyoshi3‑14‑1,\\ Kohoku‑ku, Yokohama 223‑8522, Japan\\
	\texttt{muramatsu@mech.keio.ac.jp} \\
    \And
	\href{https://www.researchgate.net/profile/Markus-Apel}{\includegraphics[scale=0.06]{orcid.pdf}\hspace{1mm}Markus Apel} \\
	Access e.V. \\ Intzestr. 5, D-52072, Aachen, Germany\\
	\texttt{m.apel@access-technology.de} \\
	\And
	\href{https://www.researchgate.net/profile/Shahed-Rezaei}{\includegraphics[scale=0.06]{orcid.pdf}\hspace{1mm}Shahed Rezaei} \\
	Access e.V. \\ Intzestr. 5, D-52072, Aachen, Germany\\
	\texttt{s.rezaei@access-technology.de} \\
}

\renewcommand{\shorttitle}{\textit{arXiv} Template}

\hypersetup{
pdftitle={A template for the arxiv style},
pdfsubject={q-bio.NC, q-bio.QM},
pdfauthor={David S.~Hippocampus, Elias D.~Striatum},
pdfkeywords={First keyword, Second keyword, More},
}

\begin{document}
\maketitle

\begin{abstract}
\begin{adjustwidth}{-1cm}{-1cm}
In this work, we introduce \textbf{implicit Finite Operator Learning (iFOL)} for the continuous and parametric solution of partial differential equations (PDEs) on arbitrary geometries. We propose a physics-informed encoder-decoder network to establish the mapping between continuous parameter and solution spaces. The decoder constructs the parametric solution field by leveraging an implicit neural field network conditioned on a latent or feature code. Instance-specific codes are derived through a PDE encoding process based on the second-order meta-learning technique. iFOL employs a purely physics-informed loss function derived via the Method of Weighted Residuals. The predicted neural field serves as the test function, resulting in the backpropagation of discrete residuals during the PDE encoding and decoding stages.\\
Compared to the state-of-the-art neural operators, iFOL introduces several key innovations: \color{black}(1) it bypasses the costly multi-network and supervised encode–process–decode pipeline of conditional neural fields for parametric PDEs; (2) it yields accurate parametric fields and solution-to-parameter gradients, enabling efficient sensitivity analysis regardless of response count; (3) it effectively captures sharp solution discontinuities, which are often challenging for some neural operator models; and (4) it is mesh and geometry agnostic, enabling zero-shot generalization to arbitrary domains. \color{black}We critically assess these features and analyze the network's ability to generalize to unseen samples across both stationary and transient PDEs. \color{black} The method is also compared against baseline operator-learning approaches, demonstrating its potential for tackling complex problems in computational mechanics.
\color{black} 

\end{adjustwidth}
\end{abstract}



\section{Introduction}
\label{sec:Introduction}
\textbf{Numerical Solvers.}
Numerical methods for solving nonlinear PDEs are fundamental to scientific computing and to the simulation of complex physical systems. Classical techniques such as the Finite Element Method (FEM), the Finite Difference Method (FDM), and Spectral Methods are widely used for their robustness, efficiency, and accuracy across different problem types. However, despite their strengths, these approaches often become computationally demanding, particularly in high-dimensional or parametric settings where the PDE must be solved repeatedly for new inputs.

\textbf{Physics-Informed Neural Networks.} Physics-Informed Neural Networks (PINN) have emerged as a promising alternative, leveraging neural networks to approximate solutions to PDEs by embedding the governing equations directly into the loss function \cite{RAISSI2019686}. In specific cases where all governing equations and boundary conditions are known, PINNs can be employed as forward solvers. Here, the term \textit{solver} refers to finding solutions that satisfy the governing equations, translating the problem into a constrained optimization task. 
\textcolor{black}{ PINN approaches also are further extended by including gradient-enhanced and modified architectures, to solve coupled PDEs in complex media \cite{ESHKOFTI2024118485, ESHKOFTI2023106908, Harandi2024}. These studies highlight the use of physics-informed loss functions, adaptive hyperparameter tuning, and the ability to accurately capture transient field variables. 
PINN has also been applied to plate bending problems, using strategies such as two-network PDE decomposition for linear variable-stiffness plates \cite{PENG2024107051} and energy-based transfer learning for geometrically nonlinear laminated plates \cite{HUANG2024118314}. Both approaches achieve high accuracy compared to FEM.   }
However, this approach faces two significant challenges: (1) the training time for PINNs is still not competitive with traditional numerical solvers, except in very high-dimensional settings where classical methods struggle \cite{REZAEI2022115616, Grossmann2024}, and (2) even at their best (considering recent advances), standard PINNs perform comparably to methods like FEM, and the solutions lack generalizability to other parametric inputs. Consequently, computational inefficiency and lack of flexibility for parametric variations persist when using PINNs solely for forward problems. The performance of PINNs in inverse problems, model discovery, and calibration differs from their application in forward problems, with several reported advantages in these contexts \cite{Gültekin2025, Faroughi2024}.

\textcolor{black}{It's worth mentioning that, as a current trend, researchers are also combining ideas from FEM and PINN to develop new deep learning-based solvers \cite{MITUSCH2021110651, skardova2024finiteelementneuralnetwork, XIONG2025117681, li2024finitepinnphysicsinformedneuralnetwork}. 
Recent FE-informed learning approaches, such as FEINNs \cite{BADIA2024116505}, combine neural networks with finite-element residuals and interpolation, demonstrating that FEM structure can significantly enhance stability and accuracy compared to standard PINN formulations. However, these methods have so far been explored mainly on 2D problems with simple geometries and homogeneous materials, without addressing parametric generalization, heterogeneous domains, or multi-resolution inference such as zero-shot super-resolution. 
\cite{LEDUC2023103904} embed an FEA-based residual into supervised learning, linking parametric inputs with structural behavior. This hybrid design improves generalization, extrapolation, and convergence compared to standard data-driven models. Parametric learning has emerged as a promising direction and is now professionally explored under the broader field of operator learning.
}


\textbf{Neural Operators.} These methods tend to learn mappings between infinite-dimensional function spaces, making them powerful tools for solving parametric PDEs and modeling complex physical systems. NOs predict solutions for varying input conditions, such as material properties or boundary conditions. 
Rather well-known approaches so far are Deep Neural Operator (DeepOnet) and its extensions \cite{Lu2021, ABUEIDDA2025117699, HE2024117130, KUMAR2025107113, yu2024separableoperatornetworks}, Physics-informed DeepOnet \cite{Wang2021, MANDL2025117586, LI2025128675}, Fourier Neural Operator (FNO) and its extensions \cite{li2021fourierneuraloperatorparametric, Azizzadenesheli2024, GEO_FNO, li2023physicsinformedneuraloperatorlearning}, Unet \cite{ronneberger2015unet, chen2019, MENDIZABAL2020101569, Mianroodi2022, GUPTA2023104709, NAJAFIKOOPAS2025110675}, \textcolor{black}{Convolutional Neural Operators (CNO) \cite{NEURIPS2023_f3c1951b}} and others.

\textcolor{black}{Despite the growing success, several well-known architectures exhibit key limitations. The vanilla version of DeepONet often suffers from spectral bias and struggles with high-frequency features or stiff responses \cite{khodakarami2025}. This has led to a growing family of DeepONet variants \cite{wang2025, LI2025128675}. FNO addresses spectral bias more effectively by learning directly in the Fourier domain. However, it is inherently limited to structured, periodic domains and requires uniform grids, often necessitating domain mapping or padding \cite{GEO_FNO}. CNO introduces a convolutional inductive bias into operator learning, offering better local feature extraction and parameter efficiency \cite{NEURIPS2023_f3c1951b}, but has been mainly applied in purely data-driven contexts on Cartesian grids. Finally, UNet-based architectures perform well on dense data with strong locality, but they typically require large parameter counts and are not inherently designed for operator learning, limiting their generalization across input functions or resolutions \cite{Mianroodi2022, NAJAFIKOOPAS2025110675}. 
See also \cite{LU2022114778} for a comprehensive comparison between different models. } 


As a result, the mentioned methods are also continuously improving their generalizability to complex geometries and their ability to handle complex solution fields \cite{Yin2024, GEO_FNO, xiao2024, zeng2025, Chen2024}.
Similar to the previous section, the combination of physics-informed NO and classical numerical solvers such as FEM, FDM, and FFT are also getting more attention, see for example \cite{Yamazaki2025},\cite{Rezaei_Najian_2025}, \cite{XU2024105714}, \cite{ESHAGHI2025117785},
\cite{lee2025finiteelementoperatornetwork},\cite{Kaewnuratchadasorn24}, \cite{Franco2023} and \cite{harandi2025}. The latter helps avoid time-consuming data generation or reduces the required data, and delivers higher accuracy for unseen predictions. 

\textcolor{black}{Recent developments such as PIFNO extend physics-informed operator learning to 1D beam problems using FNO backbones and label-free residuals \cite{VO2025121084}. 
Moreover, the variational physics-informed neural operator (VINO) is also proposed based on incorporating the variational form of PDEs directly into the operator-learning framework, enabling data-free parametric solution prediction. VINO demonstrates mesh-convergent, robust performance for smooth nonlinearities and efficiently handles standard boundary conditions on structured domains \cite{ESHAGHI2025117785}. However, its application to problems lacking well-defined variational forms, as well as to 3D nonlinear problems with complex geometries or transient PDEs, remains to be explored.} 

\color{black}
Operator learning faces several key challenges that limit its adoption in scientific computing. Models often struggle to generalize beyond training data, particularly in highly nonlinear systems, and capturing high-frequency or discontinuous solution features remains difficult, yet critical for engineering applications such as stress or flux predictions. Purely data-driven methods require large, costly datasets and frequently fail to enforce physical consistency, while extending existing frameworks to 3D problems with arbitrary geometries is still computationally demanding. 

\color{black}

\textbf{Implicit Neural Representations}
Implicit Neural Representations (INRs), also known as neural fields, represent a novel approach for encoding data as continuous functions parameterized by neural networks. Unlike traditional data storage methods, such as grids, meshes, or point clouds, INRs implicitly map input coordinates (e.g., spatial or temporal points) to corresponding output values (e.g., deformation, temperature, or color). This method enables efficient and flexible modeling of high-dimensional and complex data. Consequently, the application of INRs in scientific computing has been receiving increasing attention.

Recent work on neural fields in scientific machine learning, particularly in computational mechanics, has accelerated. \citet{serrano2023operator} proposed coordinate-based networks for solving PDEs on general geometries, removing mesh dependencies. \citet{naour2024timeseriescontinuousmodeling} used implicit neural representations and meta-learning for continuous-time modeling in time series. \citet{boudec2024learningneuralsolverparametric} introduced a physics-informed iterative solver that stabilizes optimization and speeds up convergence for parametric PDEs. \citet{yeom2025fasttrainingsinusoidalneural} achieved faster training via sinusoidal field weight scaling. \citet{hagnberger2024} developed attention-based conditional neural fields (CNF) for efficient inference, zero-shot super-resolution, and PDE generalization. \citet{Du2024} proposed a latent diffusion model using neural fields for simulating 3D turbulent flows. \citet{dupont2022coin++} introduced a neural compression method that significantly reduces encoding time via implicit representations. \citet{catalani2024neural} presented a method using INRs for steady-state fluid flow on unstructured 3D domains with shape generalization.
\textcolor{black}{
\cite{NEURIPS2024_8c2de415} proposed a space-time continuous CNF-based framework that preserves geometric information in the latent space, resulting in improved generalization and data efficiency.
\cite{guo2025} investigated CNFs for mesh-agnostic dimensionality reduction of turbulent flows, showing that domain decomposition and appropriate conditioning mechanisms improve generalization. \cite{kim2024} proposed a CNF framework for reduced-order modeling of parametric PDEs, which incorporates a physics-informed loss and exact IC/BC enforcement. \cite{hagnberger2024} proposed vectorized CNF for time-dependent PDEs, using attention-based neural fields to compute multiple spatio-temporal query points in parallel to achieve improved generalization.
Despite these advances, the potential of meta-learning \cite{vettoruzzo2024advances}, or learning to learn, for the conditional learning of discontinuous fields on irregular meshes remains largely unexplored.}

\color{black}
\textbf{Summary, open questions and our contributions.} 
Direct numerical solvers, such as FEM, are robust but require recomputation when problem parameters change, making parametric studies expensive. Neural operators offer a reusable mapping from input to solution spaces; however, most existing architectures are data-hungry and struggle with extrapolation and sharp features. Moreover, many standard neural operators have difficulty handling complex 2D and 3D geometries. Incorporating physics into these models improves generalization and stability, yet training remains challenging for diverse inputs and highly heterogeneous domains.
Conditional neural fields offer a compact, continuous representation of solution manifolds over arbitrary domains, supporting high-resolution evaluation and efficient memory usage. Recent trends in CNFs include Fourier or sinusoidal feature embeddings, and meta-learning for instance-specific codes. However, standard data-driven CNFs often rely on multi-network encode-process-decode pipelines \citep{serrano2023operator, catalani2024neural} and supervised learning. 

\textcolor{black}{\textbf{iFOL} iFOL fuses finite element concepts with conditional neural fields and meta-learning to learn continuous operators for parametric PDEs using an AD-free physics-informed (and thus unsupervised) loss function. Compared to state-of-the-art neural operators, iFOL offers (i) a streamlined architecture that avoids the costly multi-network encode–process–decode pipeline, (ii) accurate parametric fields with access to solution-to-parameter gradients for efficient sensitivity analysis, (iii) robust handling of sharp solution discontinuities, and (iv) mesh- and geometry-agnostic zero-shot generalization. These capabilities are demonstrated on stationary and transient PDEs and compared against leading operator-learning baselines.}

\color{black}

In Section 2, we summarize the main problem formulation and the types of PDEs and geometries chosen for this study. Next, in Section 3, we introduce implicit Finite Operator Learning as a promising approach for near-real-world problems. In Section 4, we present the study results, followed by conclusions and future directions in Section 5.

\section{Parameterized PDEs}
\label{sec:Parameterized_PDEs}
Let \( u: \mathcal{T} \times \Omega \times \mathcal{C} \to \mathbb{R}^d \) be the solution field representing physical quantities such as temperature, displacement, velocity, or pressure. The solution field \( u \) depends on the time \( t \in \mathcal{T} \subset \mathbb{R} \), the spatial coordinates \( {x} \in \Omega \subset \mathbb{R}^d \), and the control parameter \( c \in \mathcal{C} \subset \mathbb{R}^{d'} \). The solution field \( u \) is parameterized by \( c \) while satisfying the following PDE:
\begin{equation}
\begin{aligned}
    &\mathcal{R} = \partial_t u(t, x; c) - f(x, u; c) = 0,  & (t, x) &\in \mathcal{T} \times \Omega, \\
    & u(0, x; c) - u_0(x; c) = 0, &(t, x) &\in \{(0, x) \mid x \in \Omega \}.
\end{aligned}
\label{eq:1}
\end{equation}
Here, \( u_0 \) denotes the initial condition, and \( f \) can be either linear or nonlinear with respect to both the solution and the control parameter, and it typically involves partial derivatives of \( u \) with respect to \( x \).

The control parameter \( c \) can influence the solution field in various ways, such as defining the initial and boundary conditions, material properties, or other system characteristics. 
In more complex scenarios, \( c \) may represent a spatially distributed field, accounting for variations in geometry or domain (e.g. material) heterogeneity.  

Standard numerical methods, such as the finite element and finite volume methods, are widely used for solving Eq.~\ref{eq:1}.  
In practical applications and design processes, they are repeatedly executed for each variation in the control parameter \( c \).  
These methods yield a numerical approximation \( u(\cdot, \cdot; c) : \mathcal{T} \times \Omega \to \mathbb{R} \) for varying \( c \in \mathcal{C} \).

After applying a spatial discretization, such as the finite element method, we approximate the solution field as:

\begin{equation}
    u(t, x; c) \approx u^h(t, x; c) = \sum_{i=1}^{M} u_i(t,c) \psi_i(x),
\end{equation}

where \( \{ \psi_i(x) \}_{i=1}^{M} \) are spatial basis functions, \( u_i(t,c) \) are the time-dependent, parameterized coefficients representing the discrete solution at spatial degrees of freedom, and \( M \) is the number of grid points. Substituting this approximation into the governing PDE and applying the Galerkin projection, we obtain the semi-discrete system:

\begin{equation}
\label{eq:discrete_res}
    \mathbf{r}(\mathbf{u},\mathbf{c}) = \mathbf{M} \frac{d \mathbf{u}}{dt} + \mathbf{K}(\mathbf{u}, \mathbf{c}) \mathbf{u} - \mathbf{g}(\mathbf{u}, \mathbf{c}) = \mathbf{0},
\end{equation}

where:
\begin{itemize}
    \item \( \mathbf{u},\mathbf{r} \in \mathbb{R}^{M} \) are the vectors of unknowns (discrete solutions at spatial nodes) and nodal residuals, respectively,
    \item \( \mathbf{M} \in \mathbb{R}^{M \times M} \) is the mass matrix, given by $M_{ij} = \int_{\Omega} \psi_i(x) \psi_j(x) \,d\Omega,$
    \item \( \mathbf{K}(\mathbf{u}, \mathbf{c}) \in \mathbb{R}^{M \times M} \) represents the nonlinear stiffness, often resulting from the spatial derivatives of \( u \). It may be dependent on both the solution as well as the problem parameters.
    \item \( \mathbf{g}(\mathbf{u}, \mathbf{c}) \) is the discrete source term,
    \item \( \mathbf{c} \) is the control vector that parametrizes the discrete system by governing any or a combination of initial and boundary conditions, material properties, domain geometry, and spatial heterogeneity.
\end{itemize}

This semi-discrete system is an ordinary differential equation (ODE) system in time, which must be further discretized using a time-stepping method. In this work, the implicit Euler method is used to approximate the time-dependent term in Eq. \ref{eq:discrete_res} with a chosen time step size \( \Delta t \). Table \ref{tab:sum} outlines the parametrized boundary value problems explored in this research for operator learning. These problems are selected from a wide range of applications, particularly in computational mechanics. The upper part of the table focuses on stationary problems, where the given PDE has no transient term or time evolution, while the lower part highlights two well-known nonlinear transient equations. For clarity, the FEM-based residual vector for each problem is specified in Table \ref{tab:fe_loss}.

\newpage

\begin{table}[H]
    \centering
    \caption{ \textcolor{black}{Summary of the parametrized PDEs and their setup for operator learning.}  }
    \footnotesize
    \renewcommand{\arraystretch}{2}
    \begin{tabular}{ l c c }
        \hline
        \textbf{Steady-state problems}  & \textbf{Mechanical problem hyper~/~linear elasticity}  & \textbf{Nonlinear diffusion} \\
        \hline
        \hline
        \textbf{PDE}  & 
        \begin{minipage}{5cm} 
            \raggedright
            \[
                \nabla \cdot \boldsymbol{P} + \boldsymbol{f} = 0,~ \boldsymbol{P} = \frac{\partial W}{\partial \boldsymbol{F}}
            \]
            \[
                W = \frac{\mu(\boldsymbol X)}{2} (\bar{I_1} - 3) - \frac{\kappa(\boldsymbol X)}{4} (J_F^2 - 1 -2\ln J_F) 
            \]
            \[
                \nabla \cdot \boldsymbol{\sigma} = 0,~\hat{\boldsymbol{\sigma}} = \boldsymbol{C} \hat{\boldsymbol{\varepsilon}},~\text{see} \ref{sec:mechanics_app}
            \]
        \end{minipage} 
        &  
        \begin{minipage}{5cm} 
            \centering
            \[
                -\nabla \cdot \boldsymbol{q} = 0
            \]
            \[
                \boldsymbol{q} = - k(\boldsymbol{X},T)~\nabla T
            \]
            \[
                k(\boldsymbol{X},T)=k_0(\boldsymbol{X})(1+2 T^4),~\text{see} \ref{sec:thermal_app}
            \]
        \end{minipage} 
        \\
        \hline
        \textbf{Operator} $\mathcal{O}$: In $\to$ Out  & $\mathcal{O}: \boldsymbol{C}(\boldsymbol{X}) \to \boldsymbol{U}(\boldsymbol{X})$,~~~$\mathcal{O}: \boldsymbol{U}_b \to \boldsymbol{U}(\boldsymbol{X})$  &  $\mathcal{O}: {k}(\boldsymbol{X}) \to {T}(\boldsymbol{X})$  \\
        \hline
        \textbf{Geometry}  & 
        \begin{minipage}{5cm}
            \centering
            \vspace{0.3cm} 
            \includegraphics[width=6cm]{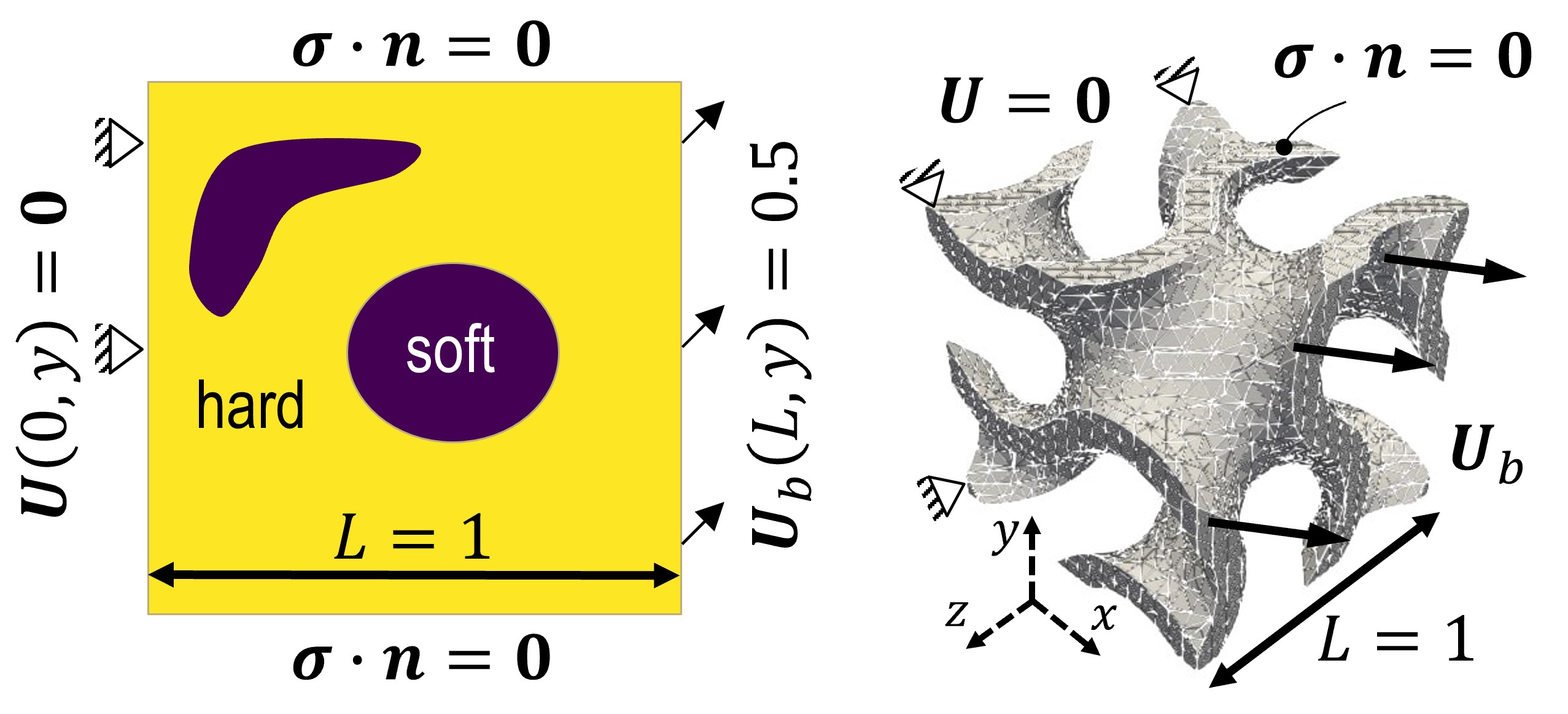} 
        \end{minipage}  
        & 
        \begin{minipage}{5cm} 
            \centering
            \vspace{0.3cm} 
            \includegraphics[width=4cm]{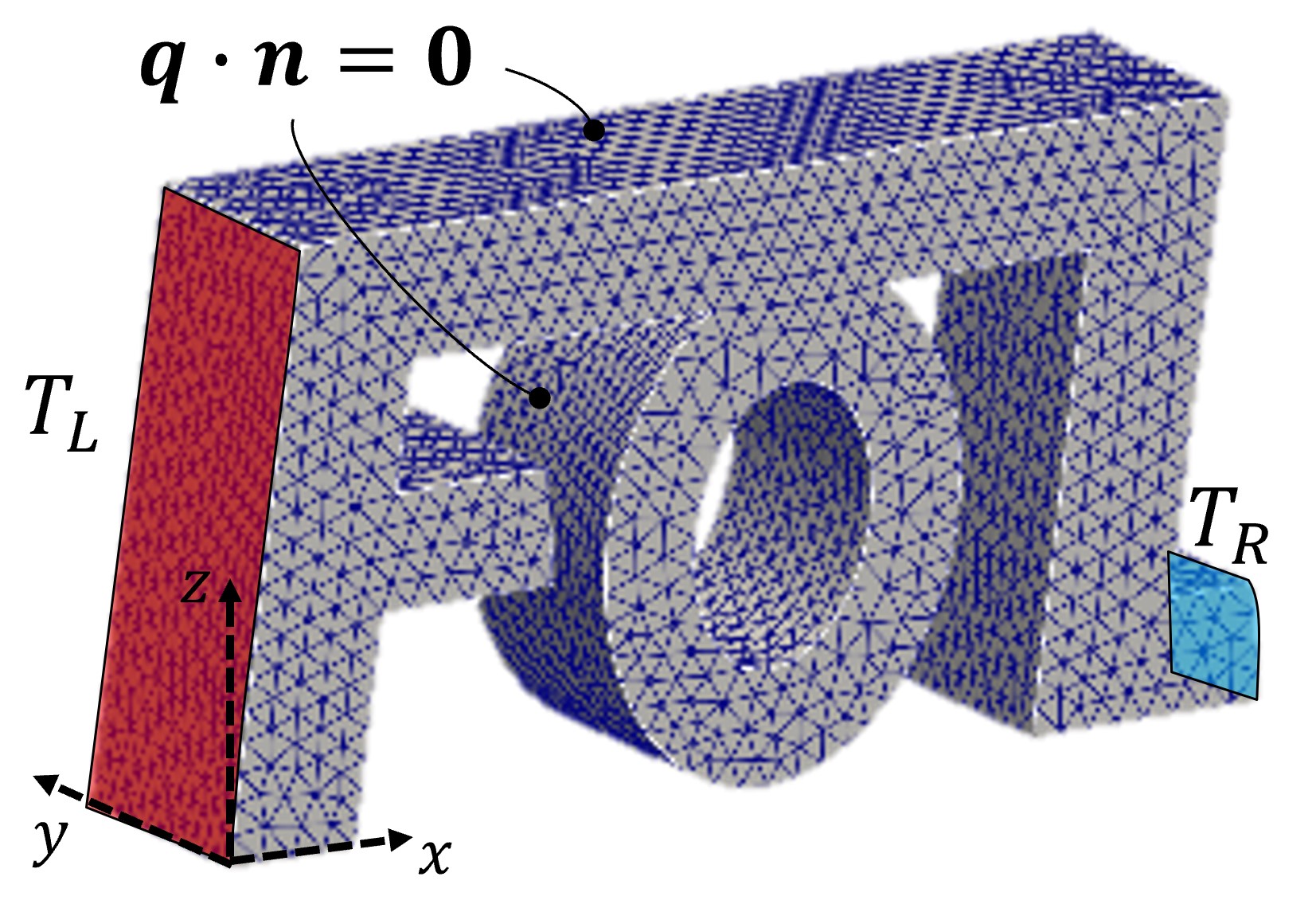} 
        \end{minipage}    
        \\
        \hline
        \textbf{BCs} &
        \begin{minipage}{5cm}
            \centering
            $\boldsymbol{U}(0,y)=0$,~~~$\boldsymbol{U}(L,y)=0.5$ \\ $\boldsymbol{U}(0,y,z)=0$,~~~$\boldsymbol{U}(L,y,z)=\boldsymbol{U}_b$
        \end{minipage}
        &
        \begin{minipage}{5cm}
            \centering
            $T_L=0.1$ on the left edge\\ $T_R=1.0$ on the right edge
        \end{minipage}
        \\
        \hline
        \textbf{Input Samples}   & 
        \begin{minipage}{5cm}
            \centering
            \[
                \boldsymbol{C}(\boldsymbol{X})= \text{Fourier-based param.}
            \]
            \[
                \boldsymbol{U}_b= \text{Random Gaussian}
            \]
            \includegraphics[width=6cm]{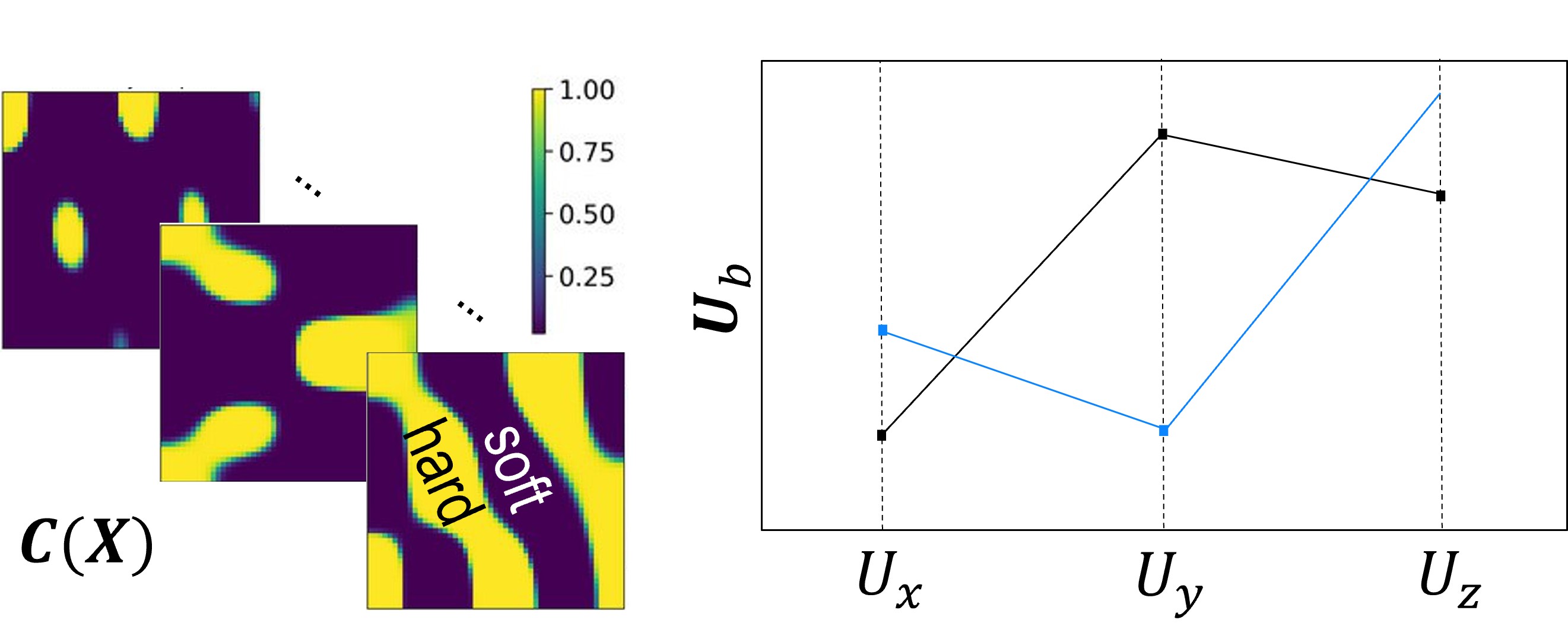} 
        \end{minipage}
        & 
        \begin{minipage}{5cm} 
            \centering
            \begin{align*}
                k_0=0.95~\text{Sig} \left(\beta \left(k_f - 0.5\right) \right) 
                + 0.05.
            \end{align*}
            \[
            k_f = \text{Fourier-based param.}, \beta=20
            \]
            \vspace{0.3cm} 
            \includegraphics[width=3cm]{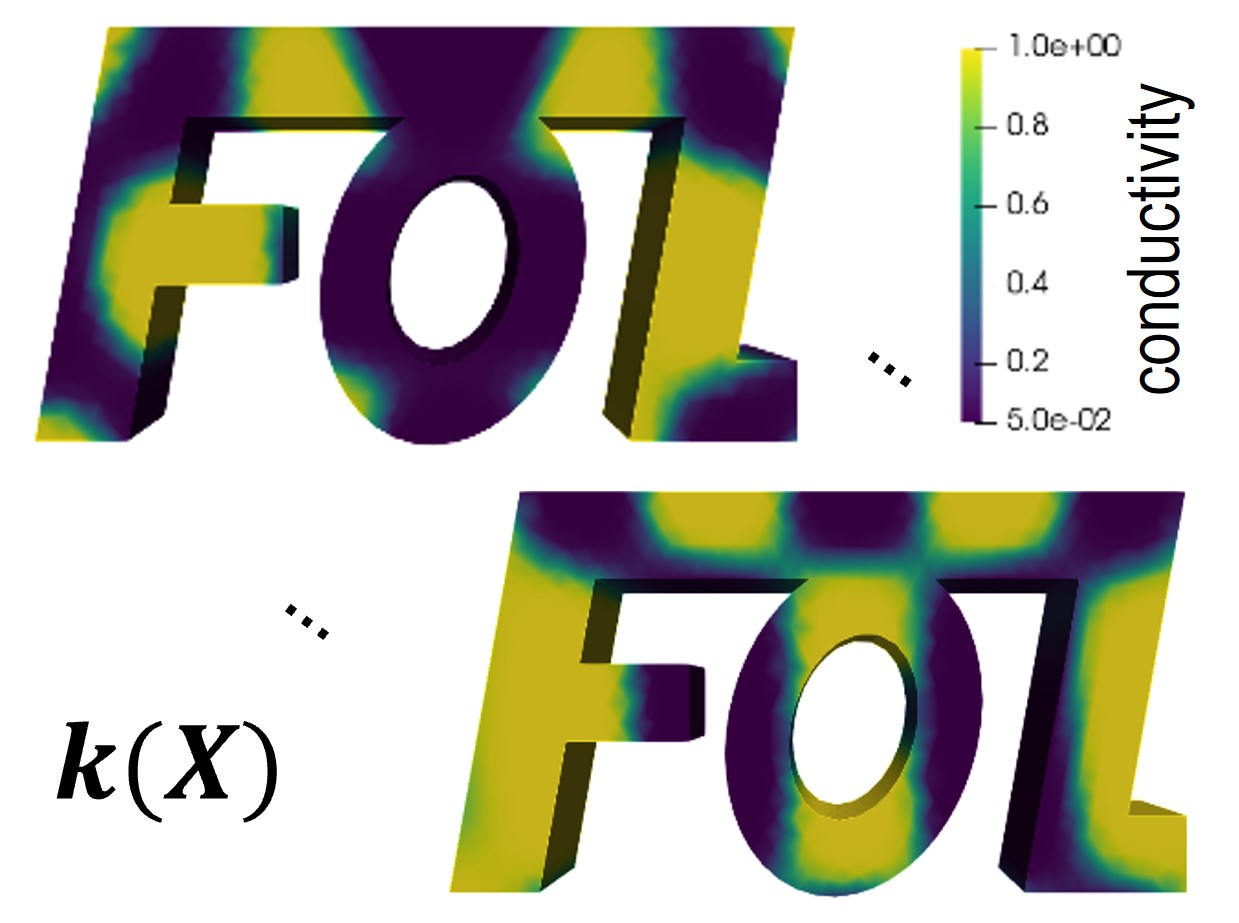} 
        \end{minipage}
        \\
        \hline
        \hline
        \textbf{Transient problems}  & \textbf{Non-linear thermal diffusion}  & \textbf{Allen-Cahn equation}  \\
        \hline
        \hline
        \textbf{PDE}  & 
        \begin{minipage}{5cm} 
            \centering
            \[
                \rho c_p \frac{\partial T}{\partial t} = -\nabla \cdot \boldsymbol{q} + Q,~\text{see} \ref{sec:thermal_app}
            \]
            \[
                \boldsymbol{q} = - K~\nabla T,~K=k_0(\boldsymbol{X})(1+\alpha T)
            \]
        \end{minipage} 
        &  
        \begin{minipage}{5cm} 
            \centering
            \[
                \frac{\partial \phi}{\partial t} = - M \left( \frac{\delta F}{\delta \phi} \right),~\text{see} \ref{sec:allen_cahn_app}
            \]
            \[
                F = \int_{\Omega} \left( \frac{\epsilon}{2} |\nabla \phi|^2 + f(\phi) \right) d\Omega
            \]
        \end{minipage}  
        \\
        \hline
        \textbf{Operator} $\mathcal{O}$, In $\to$ Out  & $\mathcal{O}: T_i(\boldsymbol{X}) \to T_{i+1}(\boldsymbol{X})$ & $\mathcal{O}: \phi_i(\boldsymbol{X}) \to \phi_{i+1}(\boldsymbol{X})$  \\
        \hline
        \textbf{Geometry}  & 
        \begin{minipage}{5cm} 
            \centering
            \vspace{0.3cm} 
            \includegraphics[width=4cm]{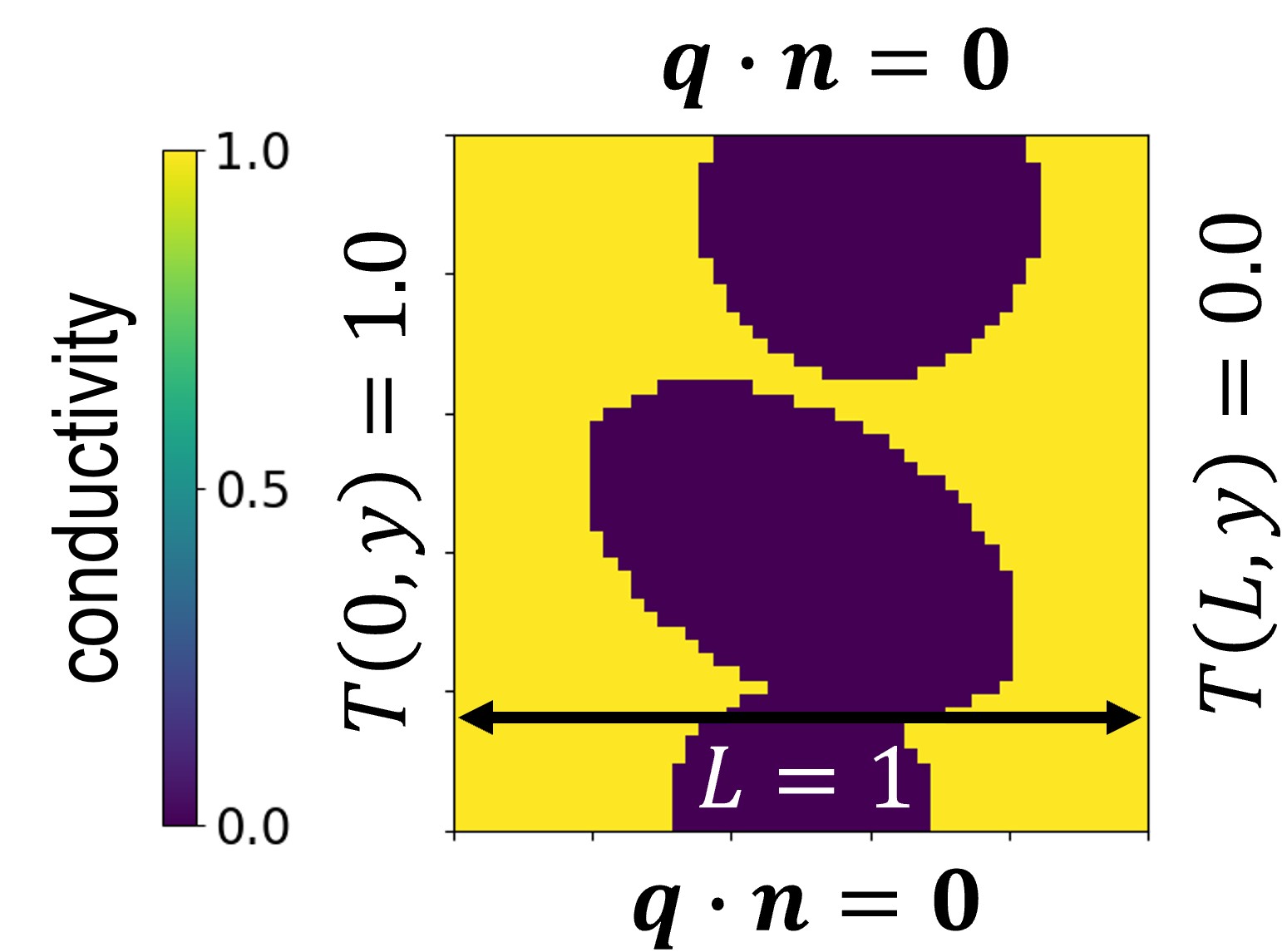} 
        \end{minipage}  
        & 
        \begin{minipage}{5cm} 
            \centering
            \vspace{0.3cm} 
            \includegraphics[width=5.5cm]{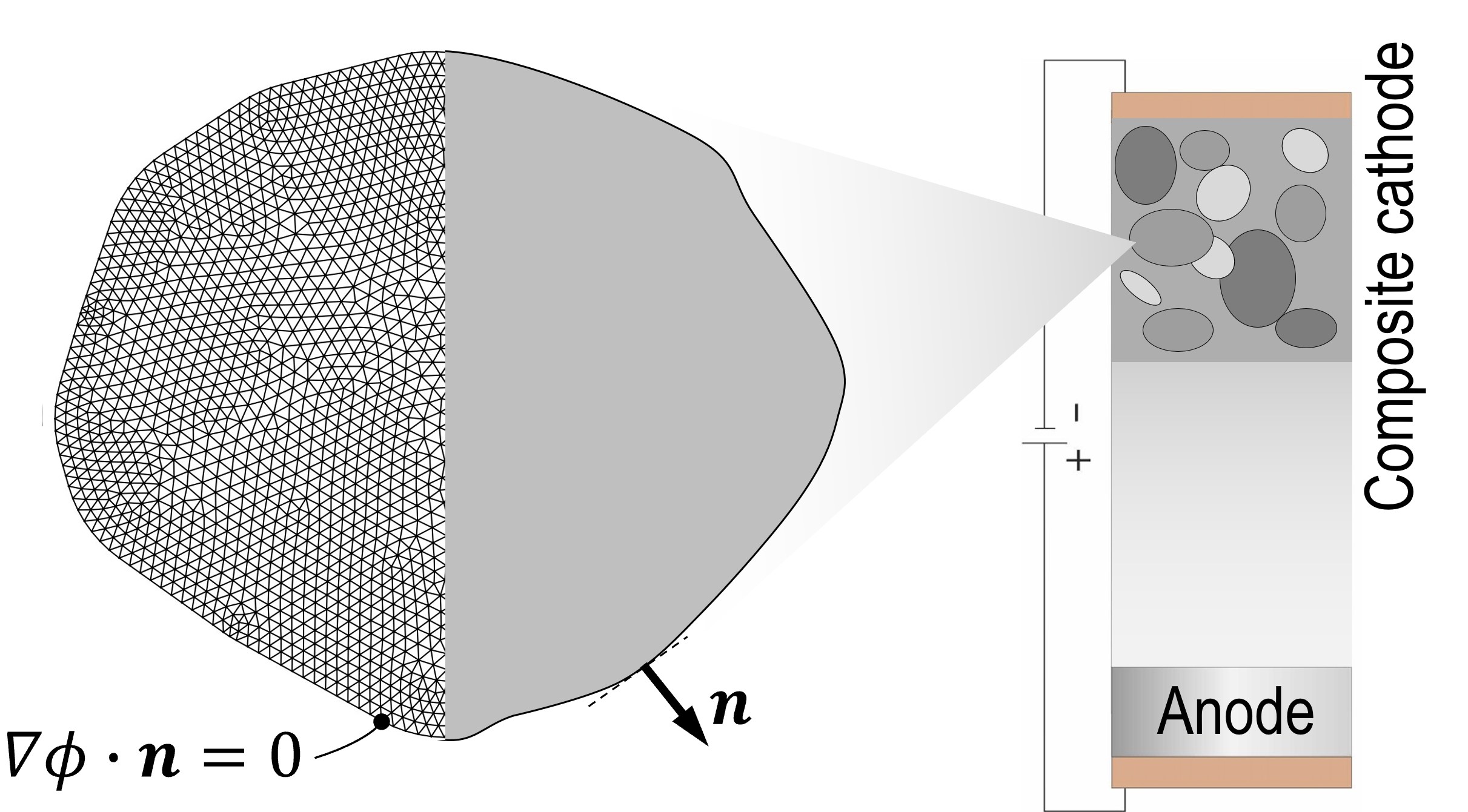} 
        \end{minipage}    
        \\
        \hline
        \textbf{BCs} &
        \begin{minipage}{5cm}
            \centering
            $T(0,y)=1.0$,~~~$T(L,y)=0.0$ \\
            $\nabla T(x,0)\cdot\boldsymbol{n}=\nabla T(x,L)\cdot\boldsymbol{n}=0.0$ 
        \end{minipage}
        &
        \begin{minipage}{5cm}
            \centering
            $\nabla \phi(x,0)\cdot\boldsymbol{n}=0.0$ at outer layer
        \end{minipage}
        \\
        \hline
        \textbf{Input Samples}   & 
        \begin{minipage}{5cm} 
            \centering
            \[
                T_0\sim \mathcal{N}\left(0, \exp\left( -\tfrac{\| X_i - X_j \|^2}{2 \varepsilon^2} \right) \right)
            \]
            \vspace{0.3cm}
            \includegraphics[width=4.5cm]{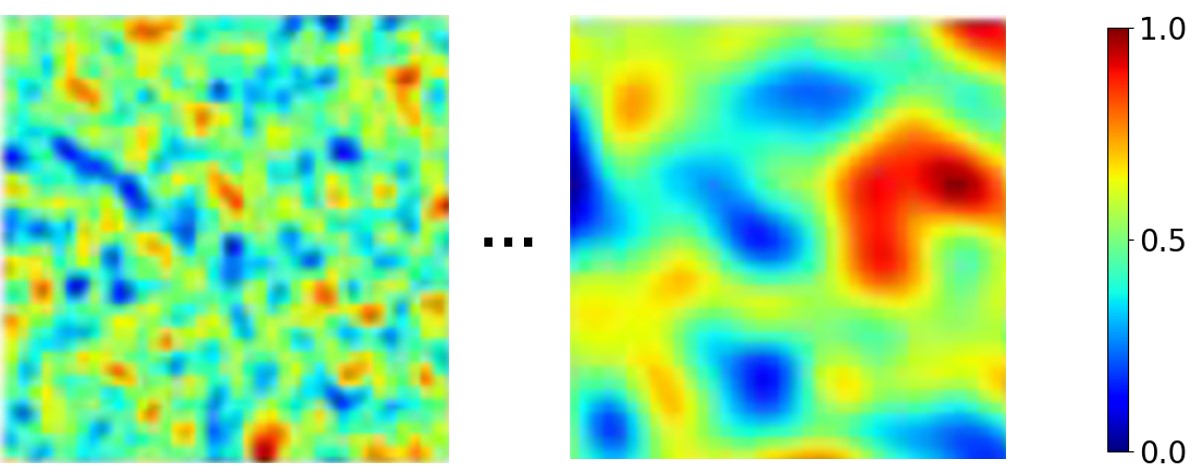} 
        \end{minipage}
        & 
        \begin{minipage}{5cm} 
            \centering
            \[ 
                \phi_0\sim \mathcal{N}\left(0, \exp\left( -\tfrac{\| X_i - X_j \|^2}{2 \varepsilon^2} \right) \right)
            \]
            \vspace{0.3cm} 
            \includegraphics[width=4.5cm]{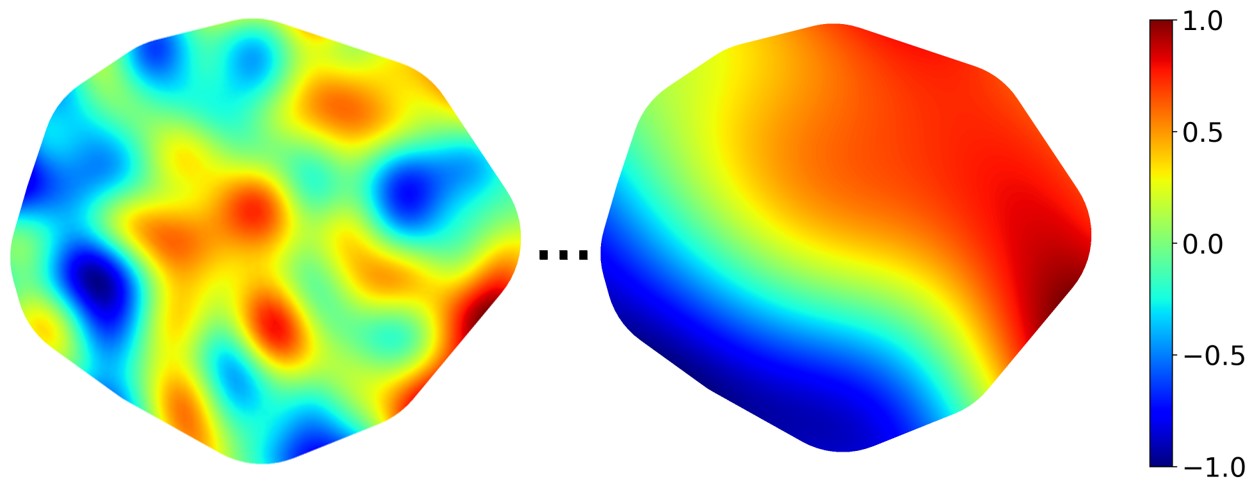} 
        \end{minipage}
        \\
        \hline
    \end{tabular}
    \label{tab:sum}
\end{table}

\newpage
\section{\textnormal{i}FOL: {i}mplicit Finite Operator Learning}
\label{sec:iFOL}

\subsection{Implicit Neural Representations}
Implicit Neural Representations (INRs) are multi-layer perceptron (MLP) networks that are coordinate-based and parameterized by \( L \) layers of weights \( \mathbf{W}_i \), biases \( \mathbf{b}_i \), and nonlinear activation functions \( \sigma_i \), with \( \theta = (\mathbf{W}_i, \mathbf{b}_i)_{i=1}^{L} \). These networks model spatial fields as an implicit function that maps spatial coordinates to scalar or vector quantities \(
x \in \mathbb{R}^d \mapsto f_{\theta}(x)\). 
We adopt SIREN \citep{sitzmann2020implicit} as the core INR architecture in our framework. This network utilizes sine activation functions, combined with a distinct initialization strategy.
\begin{equation}
\mathbf{SIREN}({x}) = \mathbf{W}_L (\sigma_{L-1} \circ \sigma_{L-2} \circ \cdots \circ \sigma_0({x}) )+ \mathbf{b}_L, \quad \text{with} \quad \sigma_i(\boldsymbol{\eta}_i) = \sin\left(\omega_0 (\mathbf{W}_i \boldsymbol{\eta}_i + \mathbf{b}_i)\right)
\end{equation}
Here, \( \boldsymbol{\eta}_0 = \boldsymbol{x} \), and \( (\boldsymbol{\eta}_i)_{i \geq 1} \) represent the hidden activations at each layer of the network. The parameter \( \omega_0 \in \mathbb{R}_+^* \) is a hyperparameter that governs the frequency bandwidth of the network. SIREN requires a specialized initialization of weights to ensure that outputs across layers adhere to a standard normal distribution.

\subsection{Conditional Neural Fields}
Our goal is to obtain a plausible solution to a parameterized partial differential equation using neural fields. This is achieved by conditioning the neural field on a set of latent variables \( \boldsymbol{l} \), which can encode the solution field across arbitrary parameterizations and discretization of the underlying PDEs. By varying these latent variables, we can effectively modulate the neural solution field. \( \boldsymbol{l} \) is typically a low-dimensional vector, and is also referred to as a feature code. See also \cite{xie2022neural}.
iFOL utilizes Feature-wise Linear Modulation (FiLM \cite{perez2018film}), which conditions the neural field in an auto-decoding manner. It employs a simple neural network without hidden layers (i.e., a linear transformation) to predict a shift vector from the latent variables \( \boldsymbol{l} \) for each layer of the neural field network. This yields the shift-modulated SIREN:

\begin{equation}
\begin{aligned}
    u_{\theta, \gamma} ({x},\boldsymbol{l}) &=\text{Decode}(\boldsymbol{l}) = \mathbf{W}_L \left( \sigma_{L-1} \circ \sigma_{L-2} \circ \cdots \circ \sigma_0 ({x}) \right) + \mathbf{b}_L, \\ 
     \sigma_i(\boldsymbol{\eta}_i,\boldsymbol{\phi}_i) & = \sin \left( \omega_0 (\mathbf{W}_i \boldsymbol{\eta}_i + \mathbf{b}_i + \boldsymbol{\phi}_i) \right), \\
     \boldsymbol{\phi}_i(\boldsymbol{l}) & = \mathbf{V}_i \boldsymbol{l} + \mathbf{c}_i.
\end{aligned}
\end{equation}

Here, \(u_{\theta, \gamma} \) is the neural solution field, \( \theta = (\mathbf{W}_i, \mathbf{b}_i)_{i=1}^{L} \) and \( \gamma = (\mathbf{V}_i, \mathbf{c}_i)_{i=1}^{L-1} \) represent the trainable parameters of the SIREN network, referred to as the Synthesizer, and the FiLM hypernetworks, referred to as the Modulator, respectively.

\subsection{PDE Loss Function}
Building on physics-informed neural networks, we introduce a domain-integrated physical loss function whose variation with respect to the solution field yields the residuals of the governing partial differential equations. Among the functionals commonly used in computational mechanics, the total potential energy functional and the weighted residual functional naturally fulfill this property. \color{black} Since the energy functional is physics-specific and not always explicitly known, we adopt the well-established Method of Weighted Residuals \citep{crandall1956engineering} to formulate the PDE loss function:
\begin{equation}
\label{eq:cont_loss}
\mathcal{L}_{\mathbf{PDE}} = \int_{\Omega} u_{\theta, \gamma} \mathcal{R} \, d\Omega = 0.
\end{equation}
The neural field \(u_{\theta, \gamma}\) serves as the test function, enforcing the vanishing of the residual in a weak (integral) sense. Method of Weighted Residuals encompasses several classical approaches, including the collocation, Galerkin, integral, and least-squares methods (see \cite{finlayson1966method} for derivations). The integral method has been widely employed in fluid mechanics; the collocation method is commonly used in chemical engineering; and the Galerkin method forms the theoretical foundation of the finite element method, which is extensively used in modern computational science and engineering.
\color{black}
Note that in the finite element method, enforcing point-wise zero residuals is justified by the the \emph{fundamental lemma of the calculus of variations} \citep{OdenReddy2011}. This lemma basically states that under sufficient regularity the only function orthogonal to all admissible test functions \(V_h\) is the zero function, i.e.,
\begin{equation}
\mathcal{L}_{\mathbf{PDE}} = \int_{\Omega} v \, \mathcal{R}(u) \, d\Omega = 0 \quad \forall v \in V_h \quad \implies \quad \mathcal{R}(u) = 0, \quad \forall x \in \Omega,
\end{equation}

In this work, the neural field $u_{\theta,\gamma}$ serves as the test function, and it satisfies $u_{\theta,\gamma}(x) \in C_c^\infty(\Omega)$ due to the use of sinusoidal activation functions.

\color{black}
To update the network parameters during backpropagation, one must compute the gradient of the loss functional with respect to the predicted solution. By invoking the chain rule in combination with variational calculus, the corresponding gradients can be expressed as follows:
\begin{equation}
\label{eq:cont_loss_variation}
\begin{aligned}
\delta_{u_{\theta, \gamma}}\mathcal{L}_{\mathbf{PDE}} & = \int_{\Omega}  \mathcal{R} \, d\Omega + \int_{\Omega} u_{\theta, \gamma} \,\delta_{u_{\theta, \gamma}}\mathcal{R} \, d\Omega  = \int_{\Omega}  \mathcal{R} \, d\Omega,
\end{aligned}
\end{equation}

Since the residual functional \(\mathcal{R}\)
should be stationary with respect to the predicted solution \({u_{\theta, \gamma}}\), the second term vanishes, leaving only the integrated residual. Thus, in continuous form, backpropagation corresponds to transporting the residuals through the network.
\color{black} To compute the loss function and residuals efficiently, we discretize the domain and employ the corresponding discrete residuals (Eq. \ref{eq:discrete_res}) as follows:  
\begin{equation}
\label{eq:PDE_loss}
    \mathcal{L}_{\mathbf{PDE}}(\mathbf{u}_{\theta, \gamma}(t,\boldsymbol{x},\boldsymbol{l}),\mathbf{c})  = \sum_{e=1}^{n_{el}}({\mathbf{u}_{\theta, \gamma}^{e}
})^T\mathbf{r}^{e},~ \text{where}~~\mathbf{r}^{e} = \mathbf{M}^{e} \frac{d \mathbf{u}_{\theta, \gamma}^{e}}{dt} + \mathbf{K}^{e} \mathbf{u}_{\theta, \gamma}^{e} - \mathbf{g}^{e}.
\end{equation}
Here, \(\mathbf{u}^e_{\theta, \gamma}\) represents the neural solution field vector evaluated at the mesh points, and the superscript \(e\) indicates that the quantity is evaluated at the element level. Here note that, although the PDE loss presented here is based on the residual, one can also directly utilize the energy functional of a PDE as shown in Table \ref{tab:fe_loss} in Appendix \ref{sec:review_FEM} as well. In fact, the energy functional-based loss in Table \ref{tab:fe_loss} was employed for the transient problems. 
\color{black} Finally, noting that \(\mathbf{r}\) denotes the nodal residual vector, the update rule for the network parameters can be derived as
\begin{equation}
\label{eq:disc_loss_variation}
\theta,\gamma \gets \theta,\gamma - \lambda \,\mathbf{r}^{T} \cdot \frac{\partial \mathbf{u}_{\theta, \gamma}}{\partial (\theta,\gamma)}, 
\end{equation}
which corresponds to the backpropagation of the discrete residuals during both the PDE encoding and decoding stages.

In cases where the governing PDEs admit a variational formulation, the user can directly minimize the corresponding energy functional, which can be more straightforward to implement compared to residual-based losses. In such cases, the loss function is defined as the discretized sum of energy contributions over all finite elements and their integration points:
$
\mathcal{L}_{\mathbf{PDE}}(\mathbf{u}_{\theta, \gamma},\mathbf{c}) = \sum_{e=1}^{n_{el}} 
\sum_{k=1}^{n_{int}} W_k^e \, \Psi\Big(\mathbf{u}_{\theta, \gamma}(\mathbf{x}_k^e), 
\nabla \mathbf{u}_{\theta, \gamma}(\mathbf{x}_k^e),\dots, \mathbf{c}\Big),
$
where $\Psi$ is the energy density at the $k$-th integration point of element $e$, $W_k^e$ are the quadrature weights, and $\nabla \mathbf{u}_{\theta, \gamma}(\mathbf{x}_k^e)$ is obtained via finite element interpolation of the neural field, without requiring automatic differentiation.

\color{black}

Remark 1: The proposed weighted residual loss function enables the seamless integration of well-established numerical methods into the learning process while minimizing computational overhead. Although the network represents a continuous solution, training is performed on a finite set of collocation points. This aligns with standard practices in numerical methods (e.g., FEM, FFT, PINNs, etc.) and neural operator frameworks, where physical constraints (or data) are enforced (or recorded) at discrete locations. \textcolor{black}{Apart from the usual approximation errors associated with finite-dimensional discretization, minimizing the integral residual rather than enforcing the strong form through pointwise constraints does not introduce additional error. Instead, the variational formulation enhances numerical stability, especially in the presence of heterogeneous coefficients, sharp features, or arbitrary geometries. See \cite{OdenReddy} for details on the mathematical proof which follows standard principle of FEM.} 


Remark 2: In transient problems, the time step $\Delta t$ is treated as a fixed parameter and directly enters the loss formulation. Here, the network essentially learns the dynamics of field evolution under a fixed time step size. While implicit integration ensures stability and accuracy for reasonable choices of $\Delta t$, we do not claim temporal super-resolution, leaving this as a direction for future improvement.

Remark 3: In this work, Dirichlet boundary conditions are strictly enforced as hard constraints by modifying the predicted solution field along the Dirichlet boundaries after inference (see also Appendix C). Neumann boundary conditions, on the other hand, are naturally satisfied through the weak formulation of the finite element method. 


\color{black}
\newpage
\subsection{Training}
The training and inference of the conditional neural fields (CNFs) involve the computation of the latent variables \( \boldsymbol{l} \), a step commonly referred to as encoding. In data-driven operator learning methodologies utilizing CNFs, encoding is performed on both the input and output fields, following the so-called Encode-Process-Decode framework\citep{dupont2022data,yin2022continuous,catalani2024neural}. In contrast, this work uniquely encodes the PDE and the underlying physics, rather than the spatial fields. In each training step of iFOL, the latent codes for the sample batch \( \mathcal{B} \) are first derived by minimizing the physical loss with respect to the latent codes in just a few steps of gradient descent:
\begin{equation}
\label{eq:encode_PDE_loss}
\begin{aligned}
\boldsymbol{l}^*(\mathbf{c}_i)&= \text{Encode} (\mathbf{PDE}) \\ &= \arg \min_{\boldsymbol{l}} \mathcal{L}_{\mathbf{PDE}} = \sum_{e=1}^{n_{el}}({\mathbf{u}_{\theta, \gamma}^{e}
})^T\mathbf{r}^{e}(\mathbf{u}^{e}_{\theta, \gamma},\mathbf{c}^{e}_i),\quad \forall i \in \mathcal{B}
\end{aligned}
\end{equation}

The parameters of the Synthesizer and Modulator networks are optimized using the computed latent codes: 
\begin{equation}
\label{eq:back_propagation_step}
\theta^*, \gamma^* = \arg \min_{\theta, \gamma} \mathcal{L}_{\mathbf{PDE}} = \sum_{e=1}^{n_{el}}({\mathbf{u}_{\theta, \gamma}^{e}
})^T\mathbf{r}^{e}(\mathbf{u}^{e}_{\theta, \gamma},\mathbf{c}^{e}_i),\quad \forall i \in \mathcal{B}
\end{equation}
Basically, we partition the model into context-specific parameters, which dynamically adapt to individual samples, and meta-trained parameters, which are globally optimized to enable knowledge transfer across diverse contexts. To the best of our knowledge, this work presents the first application of second-order meta-learning, inspired by the CAVIA algorithm \citep{zintgraf2019fast}, for the parametric learning of PDEs in a physics-informed manner. 
\color{black}
At each training step, the latent codes for a batch are obtained by minimizing the PDE loss function with respect to the codes over a few optimization steps. The resulting optimized codes are then used in updating the iFOL's network parameters $(\theta, \gamma)$ during the backpropagation. This approach is referred to as second-order meta-learning, as it implicitly involves the use of second-order derivatives of the loss function during training.\color{black}
The details of the approach are provided in Algorithms 1 and 2. Here, \( \alpha \) denotes the encoding learning rate, while \( \lambda \) is the training learning rate, which adjusts the weights of the modulator and synthesizer networks. \color{black} Based on both our experience and evidence from the literature, performing one to three gradient descent steps during the PDE encoding step is sufficient to achieve accurate and reliable performance.\color{black}

\begin{algorithm}[h]  
\caption{Encode PDE for sample batch $\mathcal{B}$} 
\label{alg:Encode_PDE}  
\begin{algorithmic}[1]  
\State \textbf{Initialize:} Set codes to zero \( \boldsymbol{l}_i \gets 0, \quad \forall i \in \mathcal{B} \);
    \For{each \( i \in \mathcal{B} \) and step \( \in \{1, \dots, K_e\} \)}
        \State \( \boldsymbol{l}_i \gets \boldsymbol{l}_i - \alpha \nabla_{\boldsymbol{l}_i} \mathcal{L}_{\mathbf{PDE}}(\mathbf{u}_{\theta, \gamma}(t,\boldsymbol{x},\boldsymbol{l}_i),\mathbf{c}_i) \) \hfill 
    \EndFor \hfill 
\end{algorithmic}  
\end{algorithm}

\begin{algorithm}[h]
\caption{Training of iFOL}
\label{alg:training_inr}
\begin{algorithmic}[1]
\While{no convergence}

    \For{\textbf{each} mini-batch \( \mathcal{M} \subseteq \mathcal{B} \)}
        \State \textbf{/* Encode PDE */}
        \For{each \( i \in \mathcal{M} \) and step \( \in \{1, \dots, K_e\} \)}
            \State \( \boldsymbol{l}_i \gets \boldsymbol{l}_i - \alpha \nabla_{\boldsymbol{l}_i} \mathcal{L}_{\mathbf{PDE}}(\mathbf{u}_{\theta, \gamma}(t,\boldsymbol{x},\boldsymbol{l}_i),\mathbf{c}_i) \) \hfill 
        \EndFor \hfill
        \State \textbf{/* Update Modulator \& Synthesizer */}
        \State \( \theta,\gamma \gets \theta,\gamma - \lambda \frac{1}{|\mathcal{M}|} \sum_{i \in \mathcal{M}} \nabla_{\theta,\gamma} \mathcal{L}_{\mathbf{PDE}}(\mathbf{u}_{\theta, \gamma}(t,\boldsymbol{x},\boldsymbol{l}^*_i),\mathbf{c}_i) \);
    \EndFor
\EndWhile
\end{algorithmic}
\end{algorithm}

\begin{figure}[H]
  \centering  \includegraphics[width=0.95\linewidth]{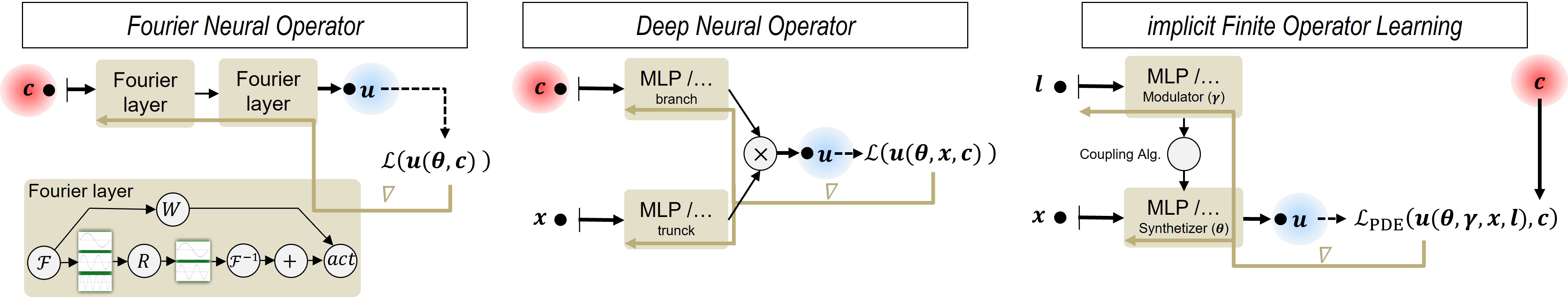}
  \caption{{Comparison of different architectures for operator learning. The brownish arrows indicate the flow of gradients in each method, while the input parametric space and output continuous field are highlighted with reddish and bluish circles, respectively. In the proposed iFOL method, the input space first influences the loss term.} }
  \label{fig:compare}
\end{figure}

\begin{figure}[H]
  \centering  \includegraphics[width=0.95\linewidth]{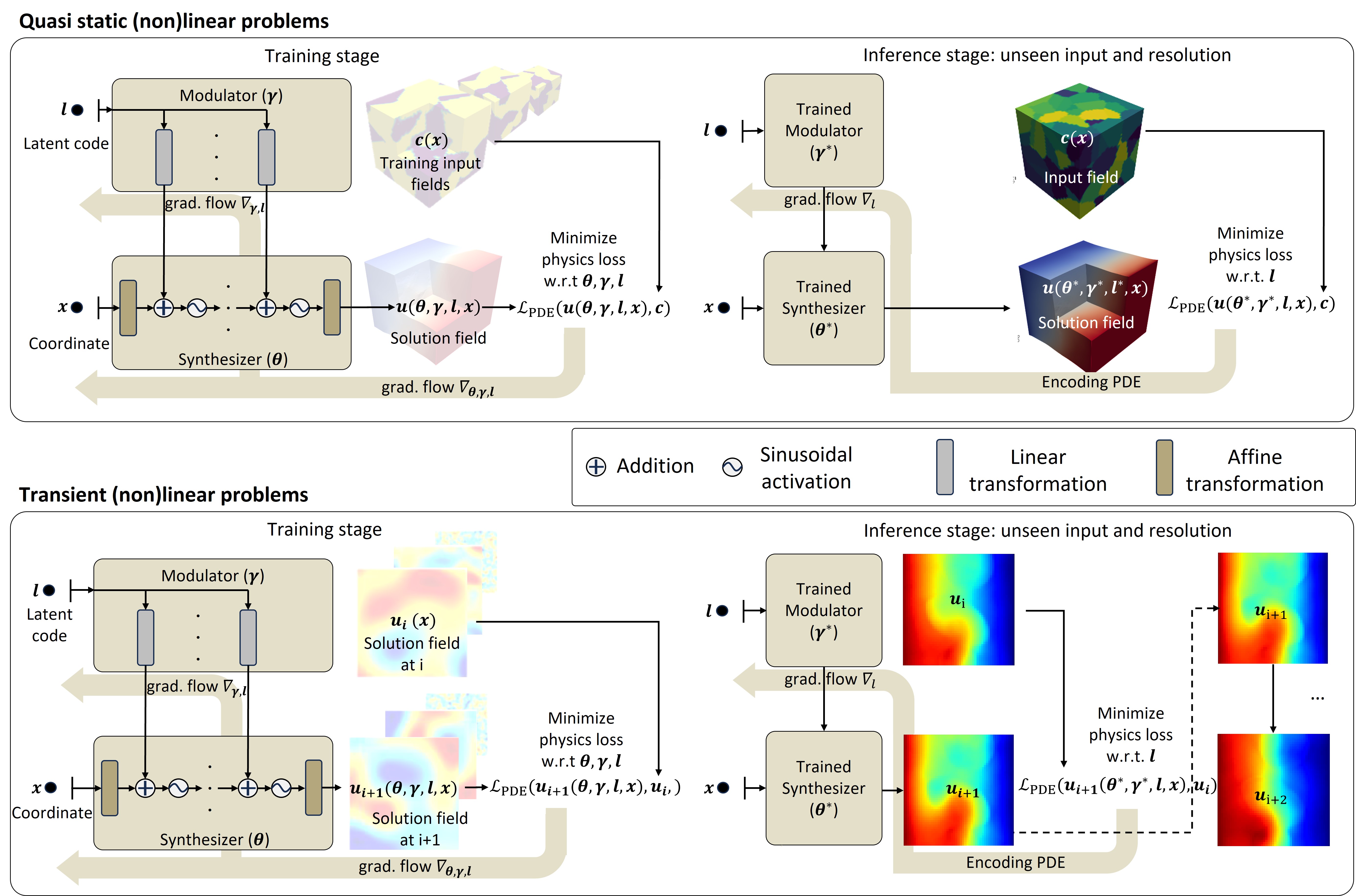}
  \caption{Training and inference procedures in the iFOL framework. Top: Details of the iFOL architecture for quasi-static problems, where the goal is to predict the corresponding solution for a given input parameter space in a single step. Bottom: Details of the iFOL architecture for transient problems, where the trained network is repeatedly called to predict the solution field over time. }
  \label{fig:iFOL_all}
\end{figure}

To provide further clarity on the architecture presented above, a comparative analysis between the proposed approach and two well-established operator learning algorithms is shown in Fig.~\ref{fig:compare}. Furthermore, a more detailed description of each component of the iFOL framework is provided in Fig.~\ref{fig:iFOL_all}. The upper section of the figure outlines the approach for stationary problems, while the lower section demonstrates the application of the same architecture to transient problems. 
\color{black}

\color{black}

In the transient examples, we adopt a physics-informed autoregressive formulation to ensure causality and properly capture the path-dependent evolution of the solution. While one could theoretically introduce time as an additional continuous input and directly predict the final state, such approaches are not easy to train for multidimensional problems and often rely on additional data to learn the patterns better and require possible enhancement regarding respecting causality. A systematic comparison between both strategies is an important direction for future work, where their respective advantages, limitations, and computational trade-offs should be examined in greater depth.

\color{black}

\section{Results}
\label{sec:Results}


\subsection{Stationary problems: hyper-elastic mechanical equilibrium}
\label{sec:mechanics}
We begin by evaluating the performance of iFOL in predicting the mechanical deformation of a hyperelastic heterogeneous solid when the property distribution varies, i.e., the operator maps \(\mathcal{O}: \boldsymbol{C}(x,y) \to \boldsymbol{u}(x,y)\). The strong form and energy formulation of the solid in this context follows a nonlinear form, which is detailed in Tables~\ref{tab:sum} and \ref{tab:fe_loss} and Section \ref{sec:mechanics_app}, along with the selected material properties.  

\begin{figure}[H]
  \centering
\includegraphics[width=0.99\linewidth]{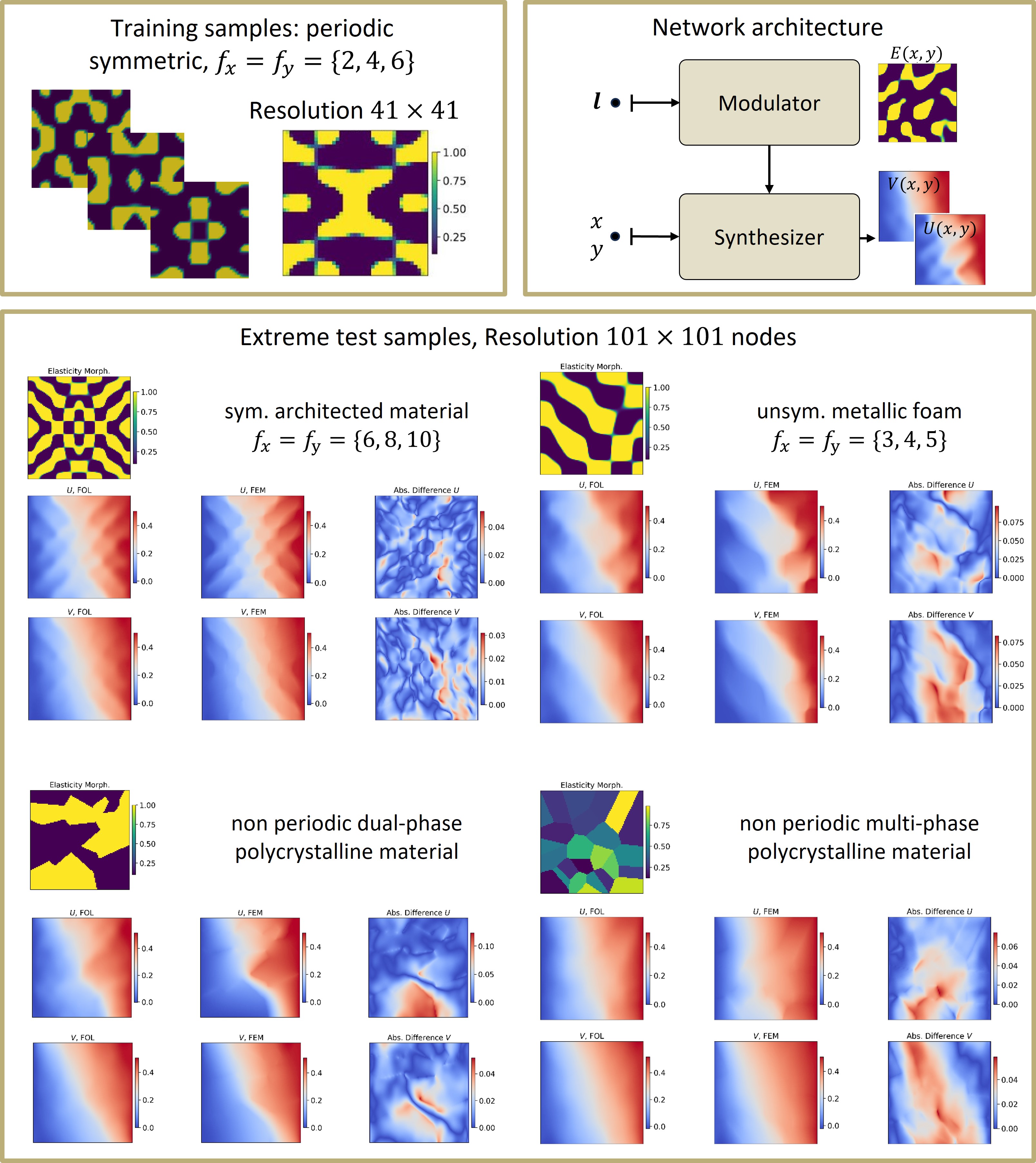}
  \caption{ Results for the 2D stationary mechanical equilibrium PDE considering a hyperelastic material model. The operator learns to map the elasticity distribution to the deformation fields (see also Table \ref{tab:sum} and Section \ref{sec:mechanics_app}). }
  \label{fig:2D_mech}
\end{figure}

\color{black}
The $8000$ random training samples are generated using a Fourier-based parametrization
The Fourier-based function for representation of the elasticity map in a 2D setting has the following form (See also \cite{rezaei2025finite} for details on this parameterization scheme):
\begin{equation}
\label{eq:foruier_eq}
\begin{aligned}
    E_f(x,y) &= \sum_i^{n_{sum}} D_i \cos{(f_{x,i}~x)} \cos({f_{y,i}~y)}].
\end{aligned}
\end{equation}
In Eq.\,\ref{eq:foruier_eq}, $D_i$ represents the amplitudes of the corresponding frequency number and $\{f_{x,i},~f_{y,i} \}$ represents the frequency in the $x$ and $y$-direction. 
Finally, we specify the property function $E_f$ through a so-called Sigmoidal projection as shown below to have more realistic and interesting characteristics for the microstructure.
\begin{equation}
\label{eq:sigmoid}
\begin{aligned}
    E(x,y) = (E_{max}-E_{min})\cdot\text{Sigmoid}\left(\beta(E_f-0.5)\right) + E_{min}. 
\end{aligned}
\end{equation}
With the above projection, we ensure that the value of Young's modulus lies between $E_{min}=0.1$ and $E_{max}=1.0$ MPa. With the $\beta$ parameter, one can also control the transition between the two phases, which is set to $\beta=20$ for the samples in this study. Some of possible samples that can be processed, are shown in Fig.~\ref{fig:2D_mech}.

We select three frequencies for each direction. Considering the constant term, this choice gives us $M = 3 \times 3 + 1 = 10$ different terms, which can be added up to construct $E(x,y)$ according to Eq.~\ref{eq:sigmoid} and Eq.~\ref{eq:foruier_eq}.
\color{black}

For this study we use \( f_x = \{2, 4, 6\} \) and \( f_y = \{2, 4, 6\} \). The chosen network parameters are listed in Table \ref{tab:hyperparam_stan}.  
In Fig.~\ref{fig:2D_mech}, the results of iFOL are shown for four unseen test cases, all evaluated at more than twice the training resolution (i.e., approximately six times more grid points). 

The first row presents predictions for unseen higher-frequency components as well as low-frequency ones with an unsymmetrical property distribution. In the second row, we further challenge the network by testing its performance on polycrystalline microstructures, which go well beyond the training samples used in this study. In particular, in the last case featuring a multiphase polycrystal, we even alter the material property values (e.g., Young's modulus) to include ranges not encountered during training.  

\textcolor{black}{Despite significant changes in input space and resolution, the network produces reasonable predictions. Across all test cases, the average maximum pointwise error remains around $0.067$, while the relative error in the averaged deformation components stays below 5\% for out-of-distribution inputs and below 1\% for in-distribution input property spaces.}

\begin{figure}[H]
  \centering
\includegraphics[width=0.99\linewidth]{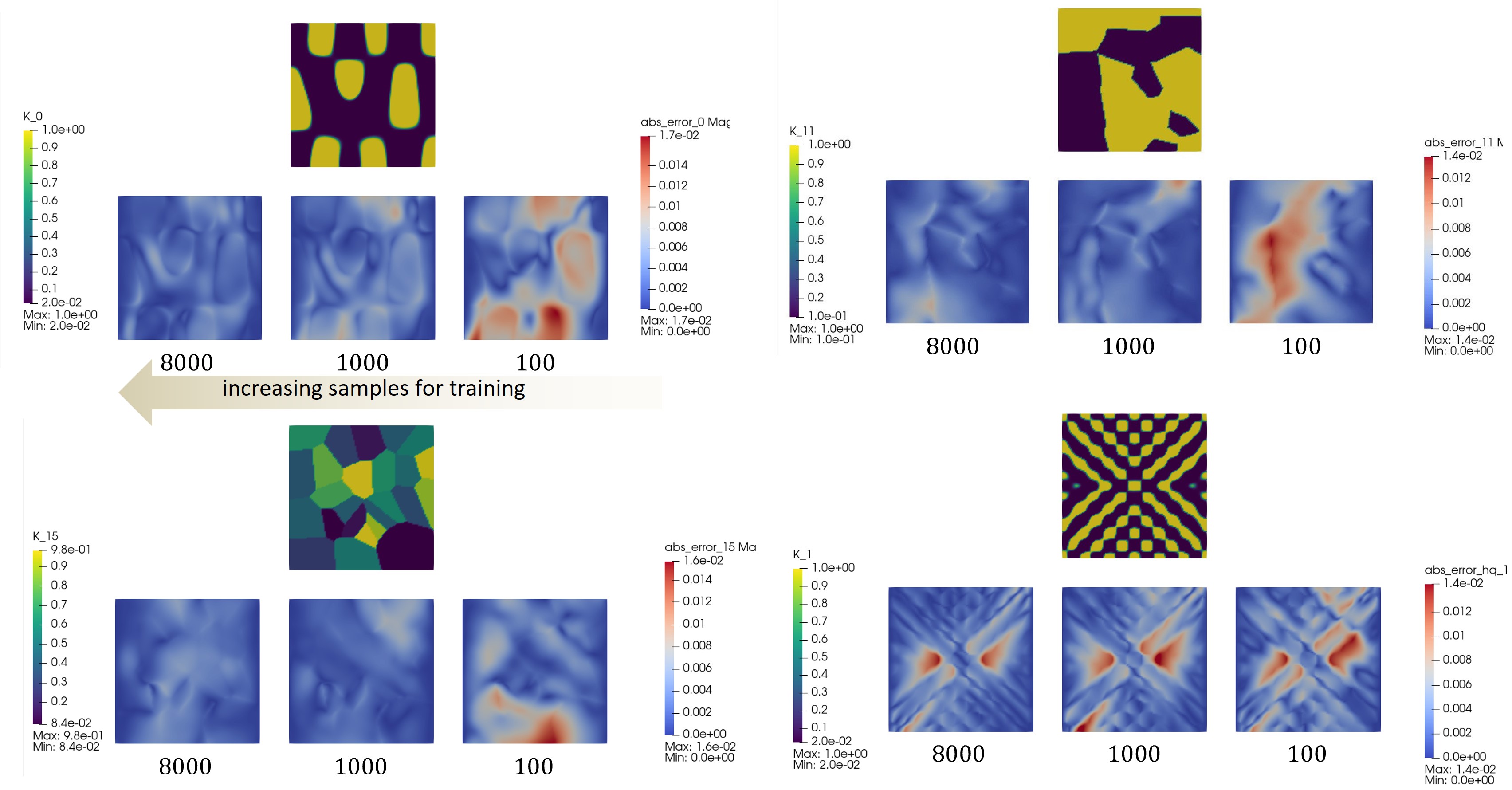}
  \caption{ Influence of the number of training samples on the error for unseen test cases and higher, untrained resolutions. Increasing the number of samples reduces the prediction errors while increasing the training time.}
  \label{fig:sample_num}
\end{figure}

At this point, a valid question is how we can ensure the quantity and quality of the initial training samples to perform a certain task. Although a solid and unique answer to this question requires much further intensive study, as it is influenced by many parameters, we attempt to address it here by altering the number of training samples while keeping the network hyperparameters the same. As expected and shown in Fig.~\ref{fig:sample_num}, increasing the number of samples systematically reduces the errors up to a certain level. Interestingly, the entire framework appears to be very sample- or data-efficient, as extensive samples are not required to achieve reasonable results. The same pattern repeats in other case studies, but we omit them for the sake of brevity.

\color{black}

Finally, we would like to further examine the strength of the proposed method in handling both sharp and smooth transitions in material properties. To this end, we performed an additional study by varying the sharpness parameter $\beta$ on a fixed, unseen test case. See also \cite{rezaei2025finite} for further details on this parametrization technique. The same trained model (without any fine-tuning) was evaluated on a higher-resolution mesh than the one used during training.

\begin{figure}[H]
  \centering  \includegraphics[width=0.85\linewidth]{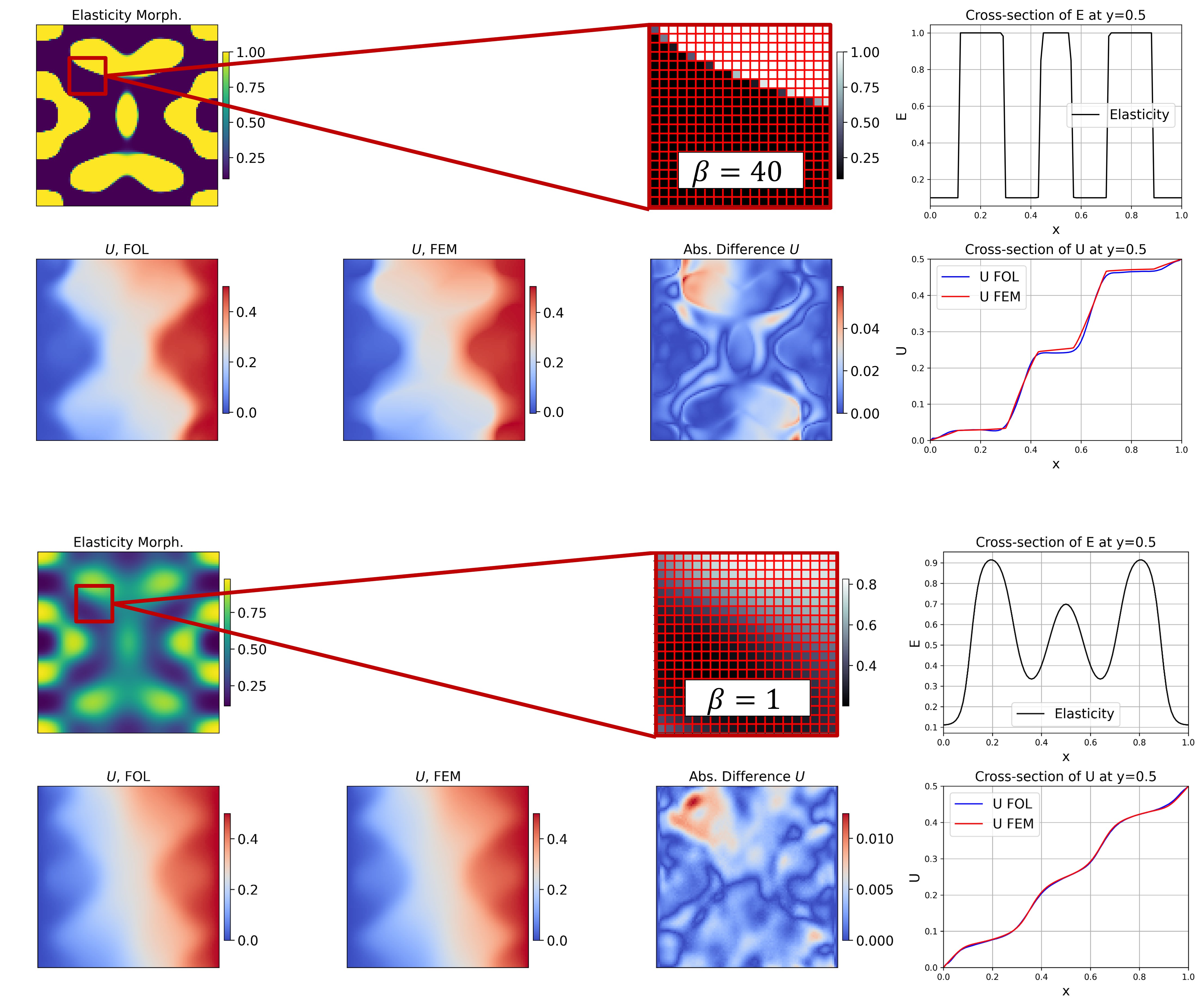}
  \caption{ \textcolor{black}{Infer iFOL at higher, untrained resolutions considering both sharply and smoothly varying elasticities.} }
  \label{fig:beta_rev}
\end{figure}

Although all training samples were generated with a sharpness parameter of $\beta=20$, the new test cases vary not only in resolution and elasticity distribution, but also in the sharpness of the transition. The first row of Fig.~\ref{fig:beta_rev} shows the model’s prediction when the sharpness is increased to $\beta=40$, while the second row demonstrates its performance for a much smoother transition with $\beta=1$, without any retraining.
We observe that for smoother transitions, the model’s performance improves, with a clear reduction in the maximum pointwise error for component $U$, decreasing from $0.032$ to $0.015$ on average across $100$ similar test cases.

\begin{figure}[H]
  \centering  \includegraphics[width=0.95\linewidth]{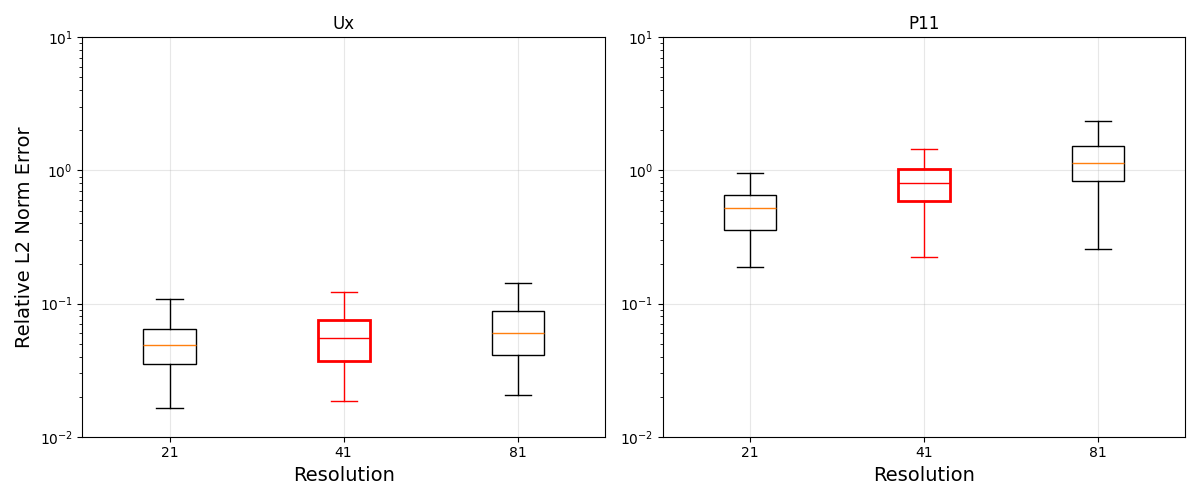}
  \caption{ \textcolor{black}{ Error analysis of the iFOL method for test samples at different resolutions. Left: errors for the primary displacement field. Right: errors for the stress component $P_{11}$. Values highlighted in red denote training-phase contrasts.} }
  \label{fig:L2_resolution}
\end{figure}

\begin{figure}[H]
  \centering  \includegraphics[width=0.95\linewidth]{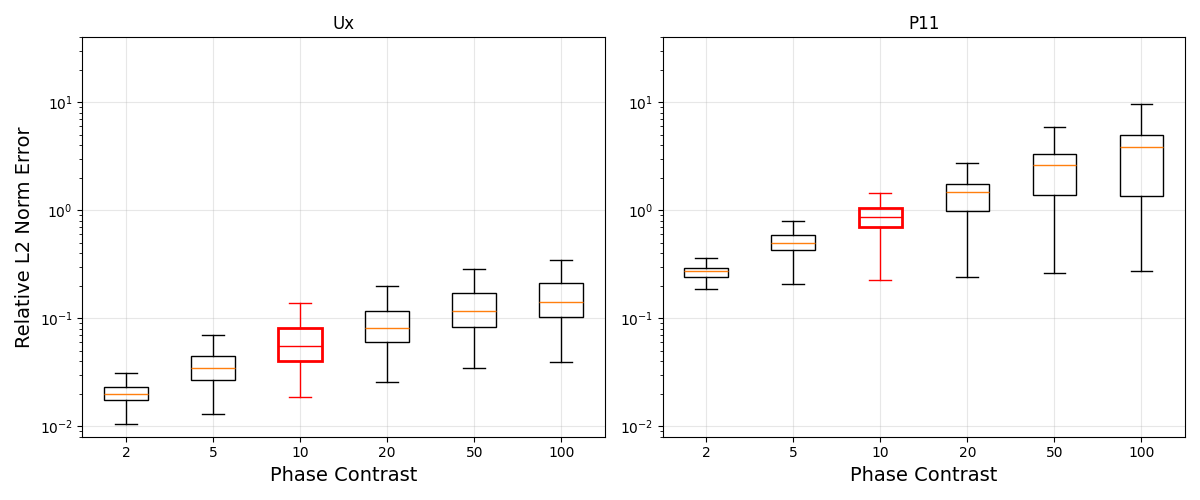}
  \caption{ \textcolor{black}{ Error analysis of the iFOL method for test samples with different phase contrast values. Left: errors for the primary displacement field. Right: errors for the stress component $P_{11}$. Values highlighted in red denote training-phase contrasts.} }
  \label{fig:L2_phase_contrast}
\end{figure}

\color{black}
The relative error norm L2 is now statistically evaluated in a wide range of out-of-distribution samples (including unseen resolutions, phase contrast values, and microstructural topologies), as illustrated in Figures \ref{fig:L2_resolution} and \ref{fig:L2_phase_contrast}. We also report the L2 norm of the predicted stress fields in separate subplots. 
In general, we observe larger errors when inferring on higher resolutions. We believe this is primarily due to material heterogeneity, which is difficult to capture accurately when the training resolution is too coarse. In such cases, a higher density of collocation points (or sensors) would be required to represent the underlying physics adequately. This also partly explains the increase in error at higher phase‐contrast values, where sharp material interfaces become more challenging for the network to approximate.
Interestingly, for lower phase‐contrast values (although never seen during training), the errors remain lower. 

Finally, we consistently observe a higher error level for the stress components. This is expected, as stress is computed from the strain field, which in turn involves spatial derivatives of the predicted solution. Consequently, these quantities are naturally more sensitive and thus contain higher errors.
If one is primarily interested in averaged (homogenized) quantities, the errors remain acceptable. However, if accurate local stresses are required, further model refinement is necessary. Possible remedies include:
1) using a neural operator that directly predicts stress (e.g., SPiFOL \cite{harandi2025}),
2) applying additional fine-tuning to reduce local oscillations \cite{taghikhani2025}, or
3) adopting ideas from mixed formulations, as in \cite{REZAEI2022115616}.

\color{black}

\subsection{Stationary problems: elasticity and learning on boundary conditions}
\label{sec:mechanics_bcs}
In this study, we examine the performance of iFOL in learning the solution for given BCs while keeping material properties and geometry fixed. More details on the chosen boundary values are provided in Section \ref{sec:mechanics_app}, while details on the formulation of the loss term can be found in Tables~\ref{tab:sum} and \ref{tab:fe_loss}. The selected network hyperparameters are listed in Table \ref{tab:hyperparam_stan}.  
Here, we define the operator as $\mathcal{O}: \boldsymbol{U}_b \to \boldsymbol{U}(x,y,z)$. 

\begin{figure}[H]
  \centering
\includegraphics[width=0.99\linewidth]{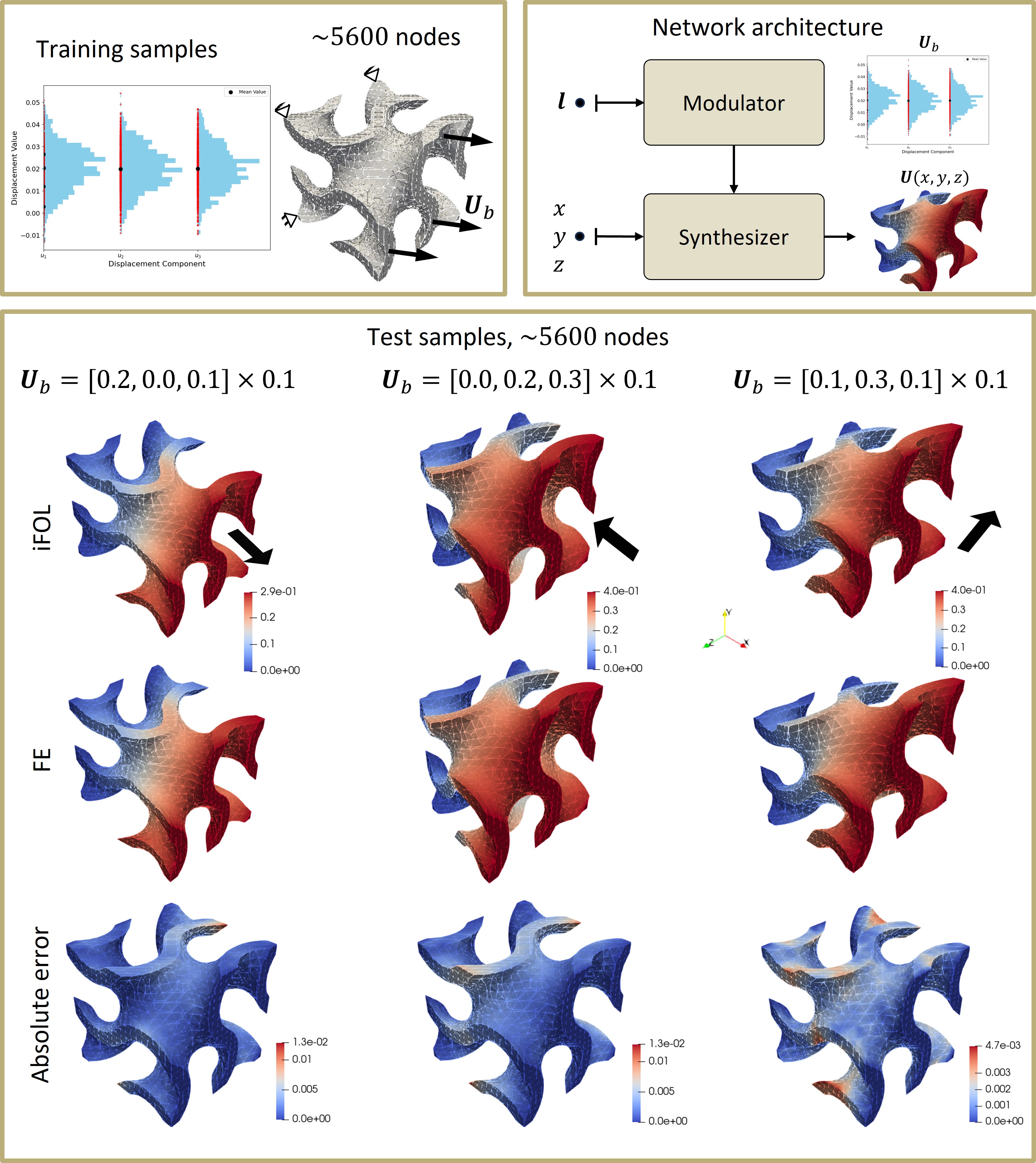}
  \caption{Results for the 3D stationary mechanical equilibrium PDE considering a linear elastic material model. The operator learns to map the given applied Drichlet boundary conditions to the deformation fields (see also Table \ref{tab:sum} and Section \ref{sec:mechanics_app}).}
  \label{fig:3D_mech}
\end{figure}

\textcolor{black}{We focus on gyroid surfaces, which are mathematically defined minimal surfaces commonly used in the design of metamaterials. Their complex topology enables novel functionalities, making them valuable in next-generation material architectures.} Moreover, we consider applications in multiscale analysis, where the applied boundary conditions typically originate from the macroscale, and it is crucial to determine the microstructural response to the imposed macroscopic displacement field.

For this purpose, we generate 1000 random samples for the applied displacement vector on the front surface of the chosen metamaterial and train iFOL to learn the full deformation field for a given Dirichlet BC. In Fig.~\ref{fig:3D_mech}, the results of iFOL versus FEM are shown for three different test cases. For visualization purposes, the deformations are magnified by a factor of 10. \textcolor{black}{Across all test cases, the average maximum pointwise error remains around $0.01$, while the relative error in the averaged deformation components stays below 5\% for out-of-distribution inputs and below 1\% for in-distribution input boundary conditions.} 
The errors are mainly concentrated in specific regions where Neumann BCs are applied. It is worth noting that, within the iFOL framework, Dirichlet BCs are enforced in a hard manner, which explains the zero error at the front and back surfaces. \textcolor{black}{Although this study focuses on Dirichlet boundary conditions, iFOL is based on FEM-inspired residuals, and therefore any boundary condition that can be handled in FEM (including mixed Dirichlet–Neumann or Robin conditions) can, in principle, be incorporated via additional residual terms or strong enforcement in the neural field.}
\subsection{Stationary problems: Nonlinear diffusion}
\label{sec:FOL_geom}
In this study, we not only critically assess the applicability of iFOL in approximating nonlinear operators on complex geometries with irregular meshes but also evaluate its effectiveness for sensitivity analysis. 
\color{black}
We are specifically interested in the derivatives of domain-integrated functionals that explicitly depend on the solution field (the operator's output) with respect to the parameter field (the operator's input), e.g.,
\begin{equation}
\mathcal{J} = \int_{\Omega} T(k)  \, d\Omega,~~~
\delta_{k}\mathcal{J} = \int_{\Omega} \delta_{k}T  \, d\Omega.
\label{eq:sa_obj}
\end{equation}

Here, $\delta_{k}\mathcal{J}$ can be computed either through adjoint-based sensitivity analysis, which serves as a reference method, or via automatic differentiation (AD) applied to the operator
 \(\mathcal{O}: {k}(x,y,z) \to {T}(x,y,z)\), which is approximated by iFOL's network and it can be computed as a postprocessing step. We also note that the computational cost of iFOL-based sensitivity
analysis is completely independent of the number of response functions, whereas the cost of adjoint-based sensitivity analysis scales directly with it.
\color{black}

For this example, 8000 samples were used for training, and 2000 samples were reserved for testing as unseen data. The conductivity field is parametrized using the Fourier parametrization as introduced in \citep{rezaei2024finite}, with $f_x=\{1, 2, 3\}$, $f_y = \{1, 2, 3\}$, and $f_z = \{0\}$. The so-called sigmoidal projection was applied to generate dual-phase conductivity fields within the range of 0.5 to 1.0. For further details, refer to the corresponding column in Table \ref{tab:sum}. A discussion and details on the network architecture are provided in Section~\ref{sec:implementation_details} and Table~\ref{tab:hyperparam_stan}.
For the summary of the mathematical formulation see also Section \ref{sec:thermal_app}. 

\begin{figure}[H]
  \centering
\includegraphics[width=0.99\linewidth]{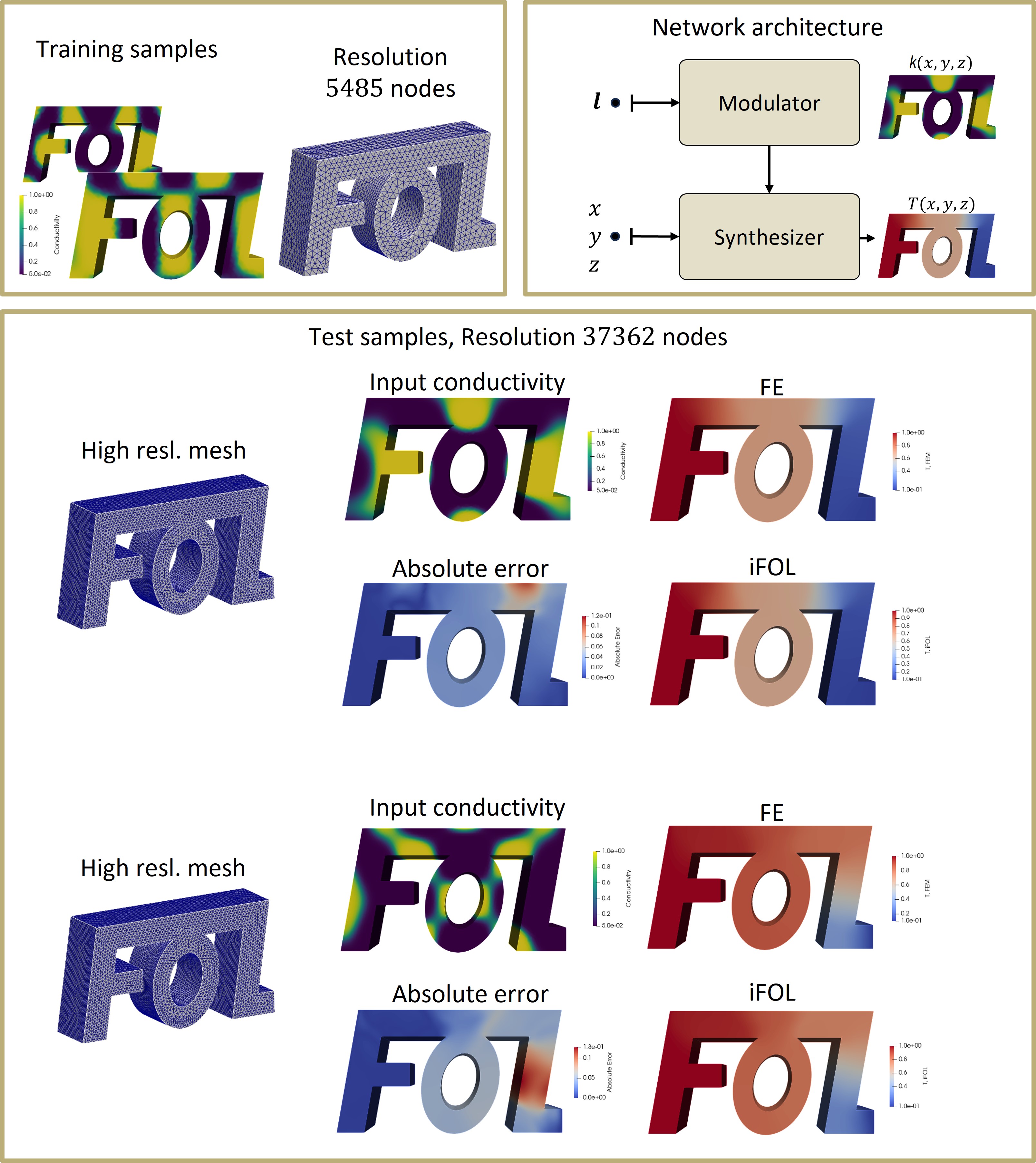}
  \caption{ Results for stationary diffusion PDE parameterized by a temperature-dependent and spatially varying conductivity field (see also Table \ref{tab:sum} and Section \ref{sec:thermal_app}). The training mesh is 7× coarser than the evaluation mesh. }
  \label{fig:3D_thermal}
\end{figure}

Figure~\ref{fig:3D_thermal} showcases how iFOL performs in predicting temperature fields for unseen conductivity distributions compared to the reference finite element method. The complicated 3D geometry of \textbf{FOL} is discretized using a fully unstructured tetrahedral mesh. The input parameter exhibits a complex spatial distribution, while the temperature field is governed by a thermal diffusion PDE with high nonlinearity arising from the temperature-dependent conductivity field. iFOL and reference FE solutions are presented for two samples that were not included in the training set. We observe that the learned operator exhibits generalization both to the parameter field and to the evaluation grid, which is 7× finer than the training mesh. 

\begin{figure}[H]
  \centering
\includegraphics[width=0.99\linewidth]{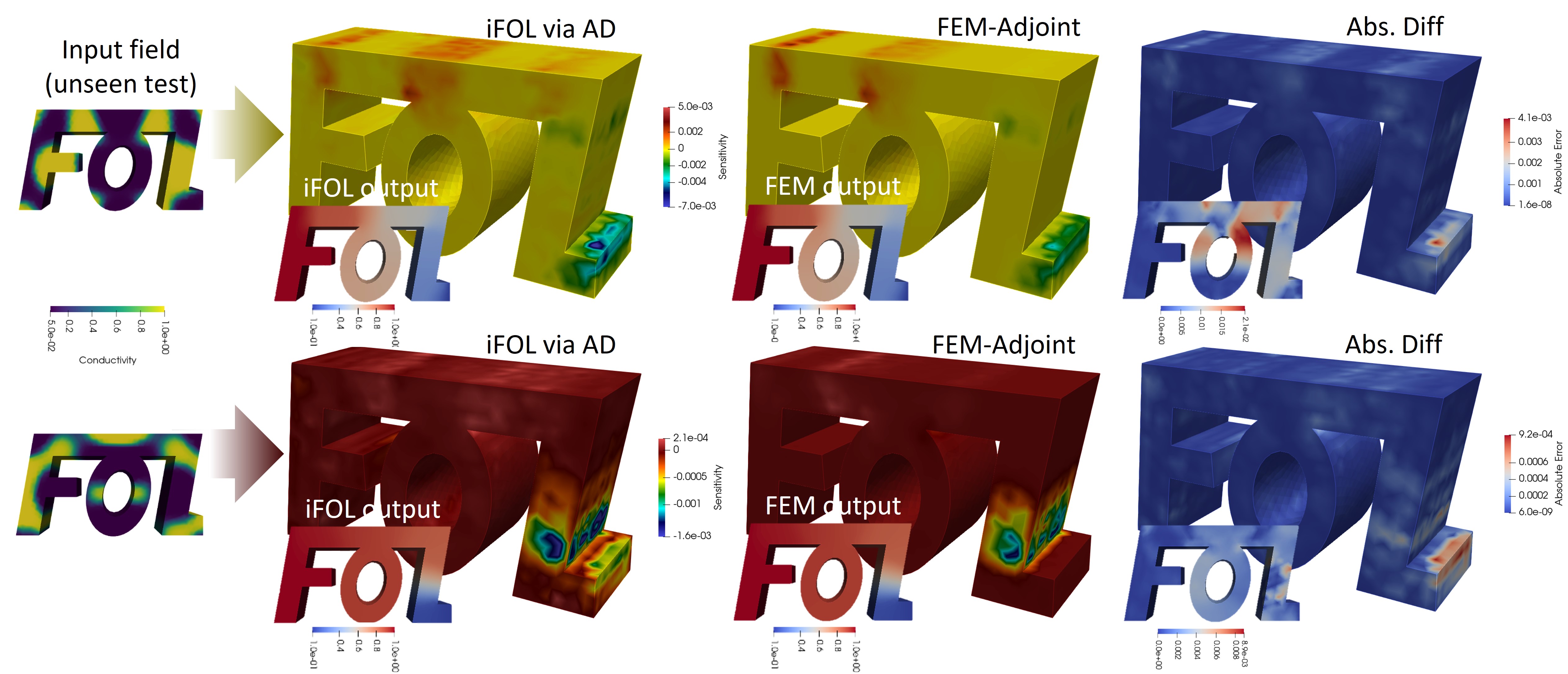}
  \caption{ \textcolor{black}{ Sensitivity maps of \(\delta_{k}\mathcal{J}\) for two unseen samples, comparing iFOL AD--based and FEM adjoint-based analysis.} }
  \label{fig:sen}
\end{figure}

Figure \ref{fig:sen} displays the sensitivity maps of \(\delta_{k}\mathcal{J}\) for two unseen samples, where one map is obtained via automatic differentiation applied to iFOL's inference function, and the other is computed using adjoint-based sensitivity analysis. Qualitative comparisons show a high degree of agreement between the two maps, with both displaying similar spatial sensitivity patterns. 
\color{black} Since sensitivity analysis and first-order optimization are mainly governed by gradient direction rather than magnitude, we investigate how well the iFOL AD--based sensitivities~$\boldsymbol{s}_1$ and the adjoint-based sensitivities~$\boldsymbol{s}_2$ agree in direction by evaluating their cosine similarity,
\[
\mathrm{cosine\ similarity}
= S_{C}(\boldsymbol{s}_1,\boldsymbol{s}_2)
:= \cos(\theta)
=
\frac{ \boldsymbol{s}_1 \cdot \boldsymbol{s}_2 }
     { \lVert \boldsymbol{s}_1 \rVert \, \lVert \boldsymbol{s}_2 \rVert },
\]
which measures the alignment of the two high-dimensional vectors. This metric captures only directional agreement, independently of 
the absolute magnitudes of the sensitivities. A value of $S_{C} = 1$ indicates perfect directional alignment, 
while $S_{C} = 0$ corresponds to orthogonality and negative values indicate opposite directions. 

For the two test cases considered, we obtain global cosine similarities 
of $0.89$ (Fig. \ref{fig:sen} top) and $0.93$ (Fig. \ref{fig:sen} down), respectively. 
These values indicate a strong directional agreement between the 
iFOL AD--based sensitivities and the FEM adjoint-based sensitivities over the full domain.
\color{black}
To the authors' knowledge, this is the first time that the Jacobian of neural network-based operators for spatially distributed input and output fields has been evaluated and compared against classical sensitivity analysis techniques, such as adjoint methods.
Based on the results presented here, we can conclude that iFOL is not only able to accurately approximate PDE-based operators in a mesh- and geometry-agnostic manner, but also to reliably estimate the operator's Jacobian.


\subsection{Transient problems: nonlinear thermal diffusion in heterogeneous domain}
\label{sec:thermal}
\begin{figure}[H]
  \centering
  \hspace*{-0.095\textwidth} 
\includegraphics[width=1.15\linewidth]{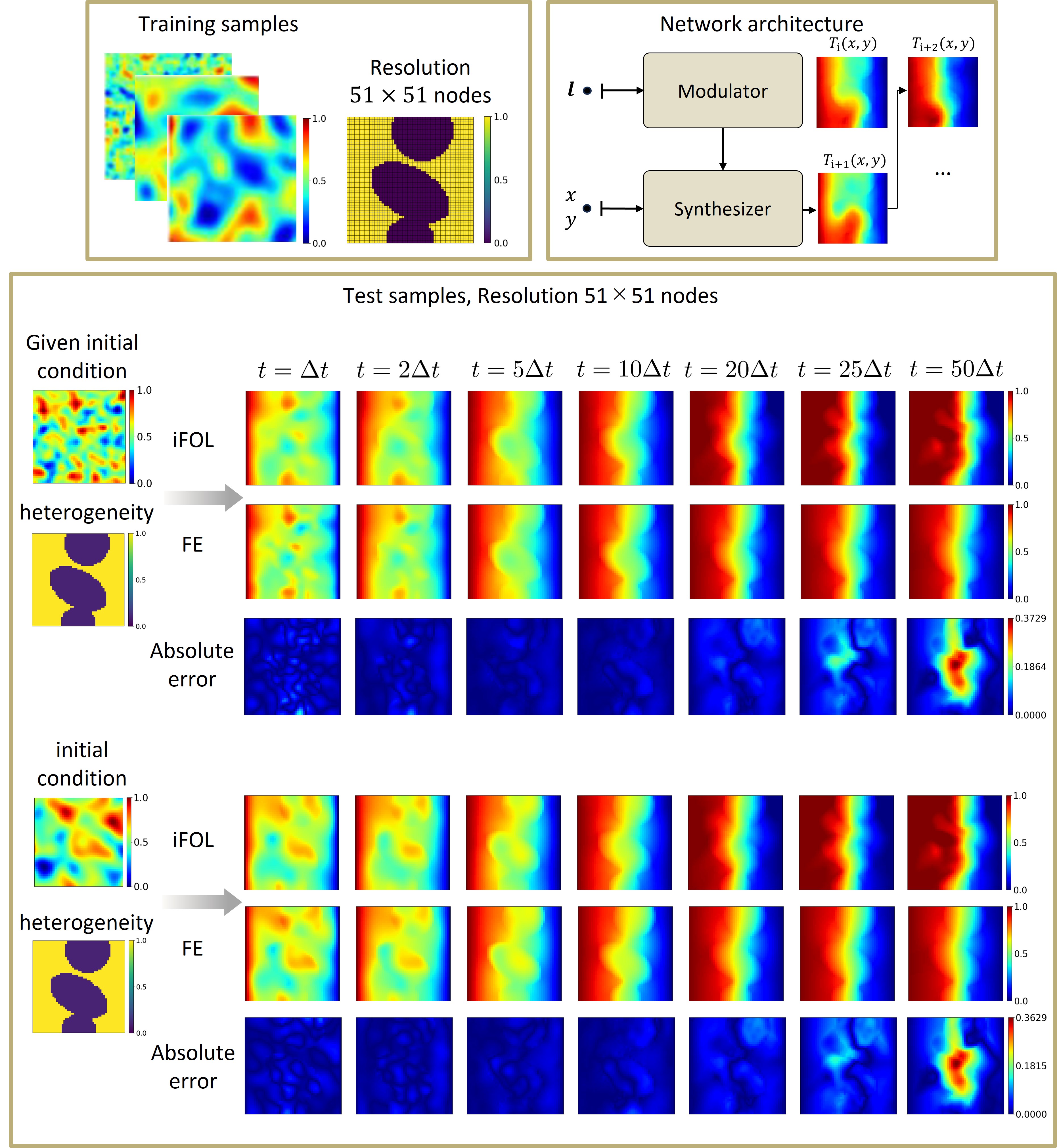}
  \caption{Results for the transient nonlinear diffusion PDE considering a temperature-dependent and heterogeneous conductivity. The operator learns to map the initial conditions to the next evolved temperature field (see also Table \ref{tab:sum} and Section \ref{sec:thermal_app}).}
  \label{fig:2D_thermal}
\end{figure}

As mentioned before, for transient problems, we adopt a slightly different strategy, where we train the network to learn the dynamics of the transient problem from step \( i \) to \( i+1 \). Upon successful training, the trained network can then be called multiple times in a loop to predict the evolution of the desired solution field. In this section, we present the results for nonlinear thermal diffusion in a heterogeneous domain where we have a map between $T_i(x,y)$ and $T_{i+1}(x,y)$. The heterogeneity is intentionally introduced to create sharp solution jumps, which typically correspond to higher frequencies in the solution and are usually not easily captured by a naive network setup. From a physical perspective, this can be related to different phases in a material microstructural system. The additional nonlinearity arises from the fact that the material conductivity is not only a function of space but also varies with temperature (i.e., the solution field). See also Section \ref{sec:thermal_app} for the summary of the mathematical formulation and further details on the material properties. The results of this study are summarized in Fig.~\ref{fig:2D_thermal}.  

Here, the training fields are randomly generated using the Gaussian process with varying length scales of the radial basis function kernel, and the results of the temperature profile evolution for different unseen initial conditions are provided. At this point, the resolution for both training and testing is the same (a grid of \( 51 \times 51 \)). Note that any obtained temperature profile as an output serves as a new input for the next step. Therefore, for transient problems, errors can accumulate over time, which has also been reported in previous studies \cite{Du2024, dingreville2024rethinking, yamazaki2025finite}. The network’s capability to handle unseen temperature fields is also worth mentioning. For example, from the fifth time step onward, the temperature profile is primarily dominated by BCs and does not resemble the randomly generated fields process used to train the network. Yet, the predictions remain reasonably accurate.  
We emphasize that the predictions of up to 50 steps shown in this figure are purely based on the one-time-trained network, with no labeled data or hybrid solver used to further support the network.  

As a final remark, we should point out our fundamentally different approach to handling transient problems, which involves breaking the time axis as explained. In the classical PINN approach, the time variable is yet another input in addition to space. Although this approach initially makes sense and appears natural, training with respect to both time and space can become very challenging for high-dimensional problems, as the network can simply violate the so-called causality, requiring additional enhancements that are not needed here \cite{WANG2024116813}.
\color{black} 
In the autoregressive approach, one-step prediction errors can inevitably accumulate over long horizons. In our formulation, the operator is trained using residual-based supervision, which penalizes the propagation of inconsistent states; however, this does not fully eliminate the risk of drift, particularly under random unseen situations not represented during training. Despite this limitation, the autoregressive structure provides a physically intuitive way to enforce causality and path-dependent behavior within nonlinear mechanical systems. 
Possible strategies to mitigate error accumulation in future developments include hybrid correction mechanisms, where neural predictions are intermittently refined using classical solvers. See investigations by \cite{LI2023116299, dingreville2024rethinking, taghikhani2025}.

\color{black}

\subsection{Transient problems: Allen-Cahn equation on the complex domain}
\label{sec:allen_cahn}
In this section, we present the results for the nonlinear Allen-Cahn equation in a homogeneous irregular domain, where we establish a mapping between \(\phi_i(x,y)\) and \(\phi_{i+1}(x,y)\). 
\begin{figure}[H]
  \centering
  \hspace*{-0.02\textwidth}
\includegraphics[width=1.05\linewidth]{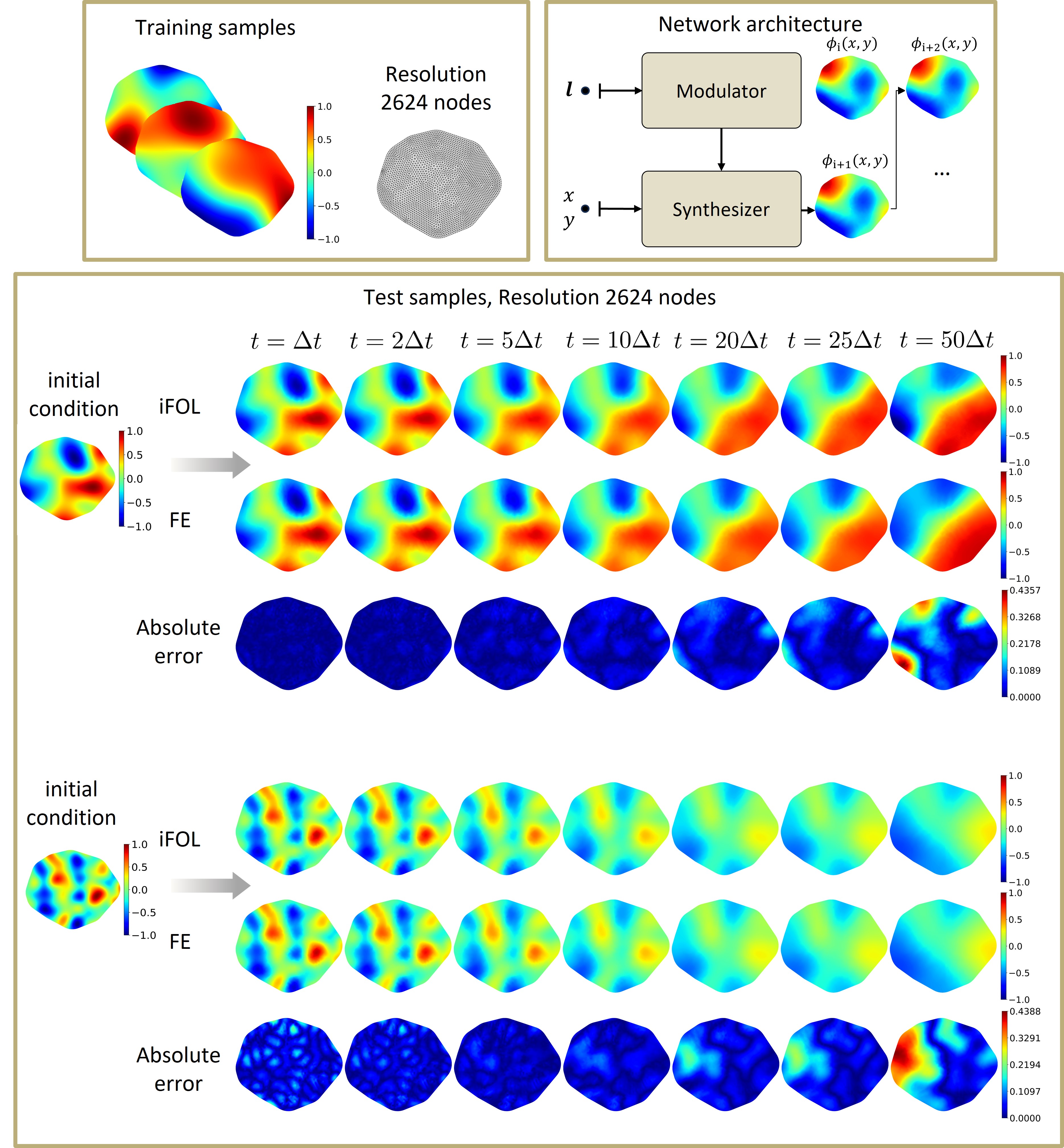}
  \caption{Results for the transient Allen-Cahn PDE considering in a complex-shape domain. The operator learns to map the initial conditions to the next evolved phase-field variable (see also Table \ref{tab:sum} and Section \ref{sec:allen_cahn_app}).}
  \label{fig:2D_AC}
\end{figure}

See also Section \ref{sec:allen_cahn_app} for the summary of the mathematical formulation.
The irregular domain is intentionally chosen to demonstrate the approach’s capability to handle complex geometries. From a practical perspective, such shapes are commonly observed in active material particles on the cathode side of battery systems, where phase transitions in the material govern ion diffusion \cite{CHEN2024234054}.  
As before, training samples are generated from the Gaussian process to train the network. The results in Fig.~\ref{fig:2D_AC} illustrate two different initializations, leading to distinct final phase profiles. The same strategy for the nonlinear thermal problem applies here, so we do not repeat every detail. Predictions remain accurate up to 10 time steps, after which errors gradually accumulate, a point that will be analyzed further.  

It is important to note that the resolution for both training and testing is kept the same, as it was found to be sufficient based on the length scale parameter in the phase field equation. Further discussion on super-resolution capabilities is provided in Section \ref{sec:zssr}.

\subsection{Studies on zero-shot super resolution}
\label{sec:zssr}
Zero-shot super-resolution (ZSSR) in operator learning refers to the ability of a trained model to enhance the resolution of function mappings without requiring additional high-resolution training data. Unlike traditional super-resolution techniques that rely on supervised learning with paired low- and high-resolution samples, zero-shot approaches leverage the inherent structure of the learned operator to generalize across different resolutions \cite{li2021fourierneuraloperatorparametric, Koopas2024}. It is worth noting the similarities to model order reduction techniques, where the dimensionality of the problem is reduced by capturing the dominant solution modes \cite{quarteroni2011reduced}.  

The ZSSR capability of iFOL is also demonstrated in Figures \ref{fig:2D_mech} and \ref{fig:3D_thermal}. Here, we further validate this claim by systematically testing the proposed iFOL approach and quantitatively assessing the error accumulation across different mesh resolutions. Our main focus is on stationary nonlinear diffusion in 3D, where we evaluate the solution on both finer and coarser grids, as well as on the 2D nonlinear Allen-Cahn equation, where we analyze error accumulation over time and for varying resolutions. Similar behavior is observed for other investigated problems, and therefore, for brevity, we do not report all details.  

\begin{figure}[H]
  \centering
\includegraphics[width=0.85\linewidth]{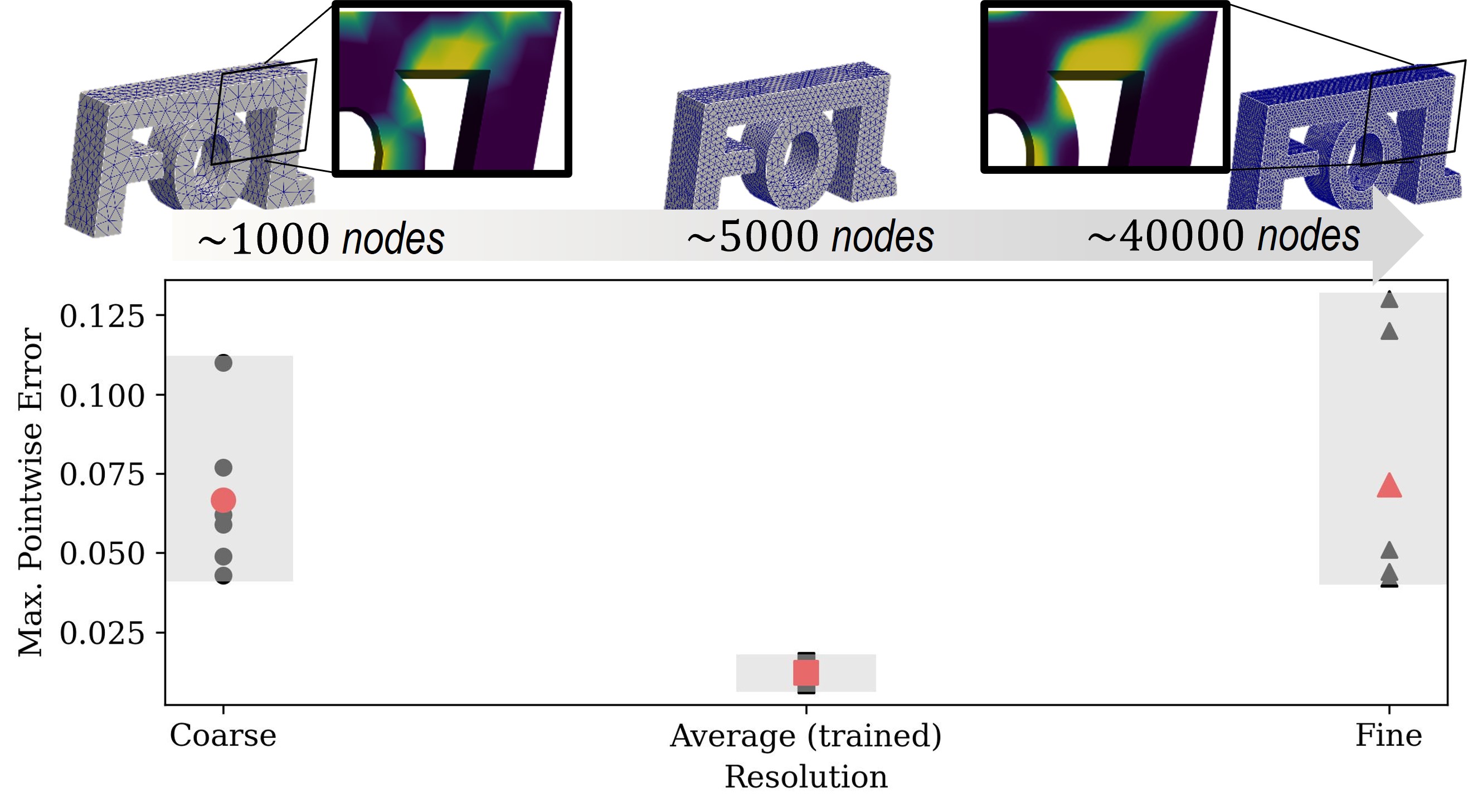}
  \caption{Performance of iFOL in ZSSR for nonlinear thermal diffusion in a heterogeneous 3D problem with unstructured meshes. The maximum pointwise error is lowest for the trained resolution and increases for other mesh resolutions, yet remains within an acceptable range. }
  \label{fig:res_3D}
\end{figure}

In Fig.\ref{fig:res_3D}, we evaluate the network on two additional mesh resolutions. Since we use unstructured meshes and nodes in the 3D domain, we aim to assess whether the network can generate results for both coarser and finer meshes. For error measurement in this particular example, we adopt a strict criterion based on the maximum pointwise error. To ensure the results are not case-dependent, we systematically test multiple samples, with five representative cases shown for each resolution in Fig.\ref{fig:res_3D}. The averaged values are highlighted in red.

As expected, the prediction error increases when moving to resolutions different from the training resolution. However, as demonstrated in Fig.~\ref{fig:3D_thermal}, the overall results remain acceptable. A potential way to mitigate this issue is to train the network on multiple resolutions across different samples, reducing bias toward a specific mesh resolution.

\begin{figure}[H]
  \centering
\includegraphics[width=0.85\linewidth]{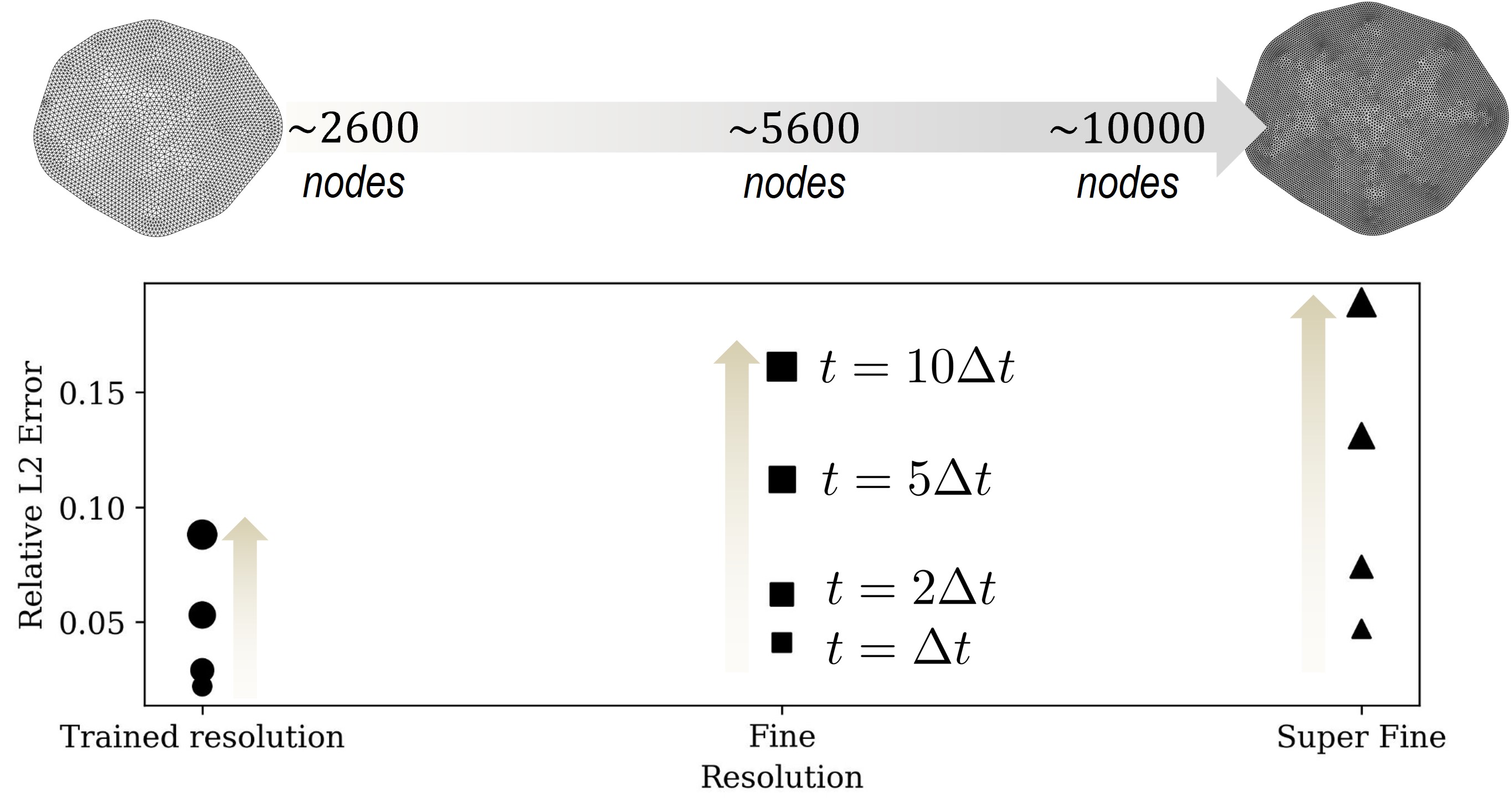}
  \caption{ Performance of iFOL in ZSSR for the nonlinear transient Allen-Cahn equation in an arbitrary 2D domain with unstructured meshes. The relative L2 error is lowest for the trained resolution but increases for other mesh resolutions and accumulates over time during further network evaluations. }
  \label{fig:res_2D}
\end{figure}

In Fig.~\ref{fig:res_2D}, we analyze the transient Allen-Cahn problem and compare results using relative L2 error measurements across three different resolutions. On the far left, we show the trained resolution, followed by two additional cases with two and four times more nodes, respectively. As expected, and similar to other physical problems, the error increases slightly when moving to different resolutions, particularly finer ones. Moreover, this diagram illustrates how errors accumulate over time for a given resolution (observe the spacing between circles, squares, and triangles along the y-axis). A potential remedy is to train the model using multiple resolutions.

\begin{figure}[H]
  \centering
\includegraphics[width=0.95\linewidth]{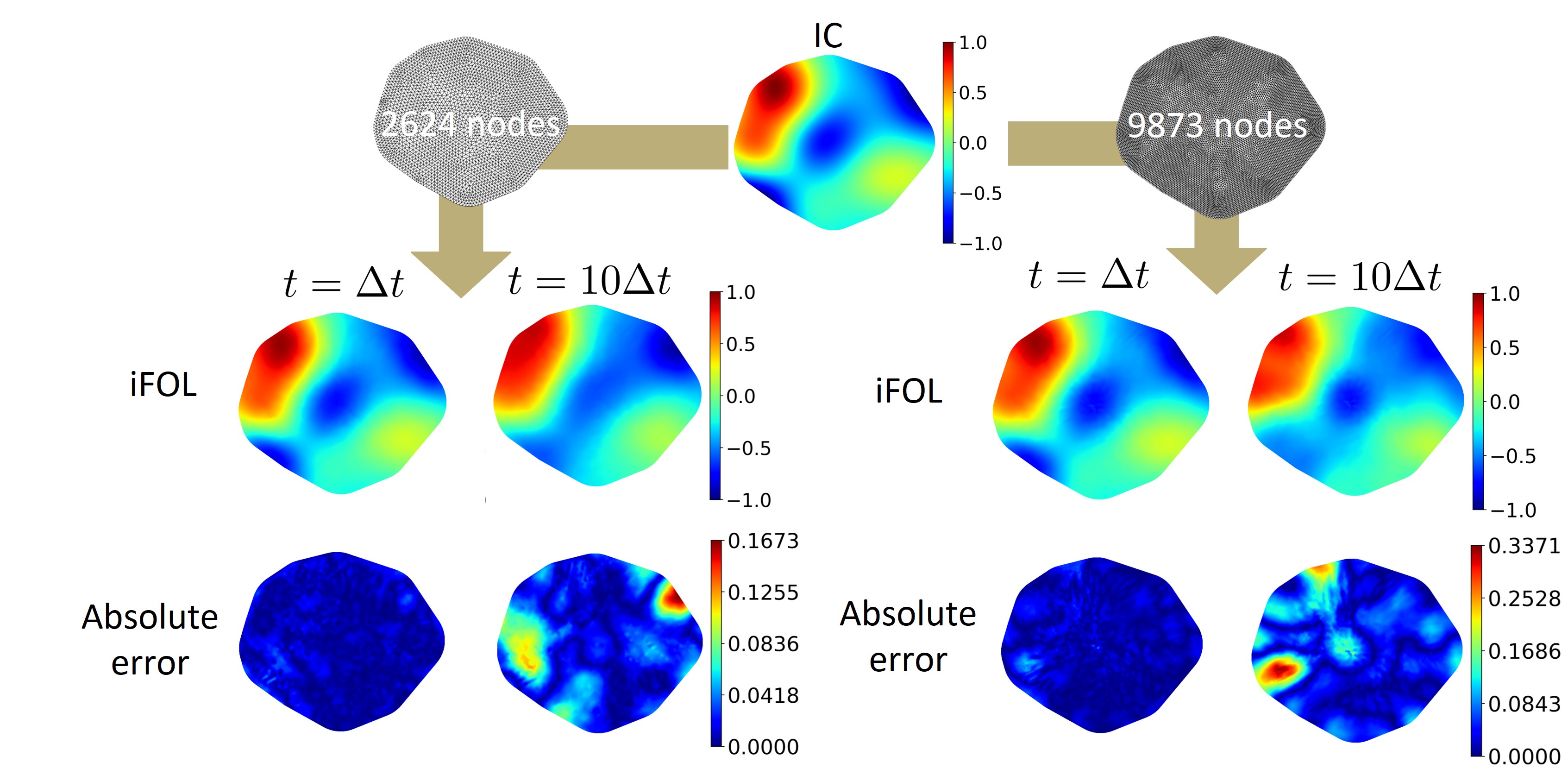}
  \caption{ Prediction of iFOL for different time steps and resolutions for the nonlinear Allen-Cahn equation.}
  \label{fig:2D_AC_ZSSR}
\end{figure}

\subsection{Computational cost analysis}
Accuracy and computational cost are two key criteria for comparing classical techniques, such as the FEM and adjoint-based sensitivity analysis, with machine learning-based approaches, including PINNs and operator learning methods. In this section, we analyze the computational cost of the iFOL technique using the 3D nonlinear diffusion problem described in Section \ref{sec:FOL_geom}. Leveraging Google JAX \cite{jax2018github}, we developed a unified framework that seamlessly integrates classical solver operations—such as element-wise computations, global Jacobian assembly, and right-hand side (RHS) construction—with machine learning techniques, including loss function evaluation and gradient-based optimization. This framework is implemented within the FOL paradigm \cite{rezaei2024finite}, utilizing key JAX features such as \texttt{jax.vmap} for efficient vectorized operations and \texttt{jax.jit} for just-in-time (JIT) compilation, enabling the translation of computationally demanding tasks into optimized machine code. The loss function is implemented exactly as formulated in Eq.~\ref{eq:PDE_loss}. Furthermore, the physical loss function not only returns the element-wise weighted residual or its energy counterpart as a scalar loss function to be minimized during the learning process but also provides the corresponding elemental Jacobian and RHS. This unified implementation enables the same framework to be used for both FEM solvers and parameterized learning approaches such as iFOL.

Table \ref{tab:network_cost} summarizes the computational times for the primal and sensitivity analyses performed using iFOL and FEM. For the sake of brevity, we focus on the example reported in Section \ref{sec:FOL_geom}, where we deal with a nonlinear PDE in a 3D complex geometry. A similar trend of results was also observed for other cases. We note that training iFOL on 8000 samples with 3800 iterations required 14 hours on NVIDIA A100 GPU with 40 GB of RAM (see Table \ref{tab:hyperparam_stan}). However, during inference, iFOL achieves speedups of 200×, 1000×, and 8300× compared to FEM on coarse (1069 nodes), normal (5485 nodes), and fine (37362 nodes) meshes, respectively. Beyond the speedup in the primal inference, iFOL also achieved 1.3× and 13× speedups for sensitivity analysis on normal and fine meshes, respectively. It should be noted that the reported times for AD-iFOL and Adjoint-FEM correspond exclusively to sensitivity analysis, assuming the primal solution is already available. Furthermore, the computational cost of iFOL-based sensitivity analysis remains independent of the number of response functions. In contrast, adjoint-based sensitivity analysis scales directly with the number of responses, as a separate adjoint system of equations must be solved for each response.

\begin{table}[H]
    \centering
    \caption{Computational costs of the studies in the 3D nonlinear diffusion problem Sec. \ref{sec:FOL_geom}.}
    \label{tab:network_cost}
    \begin{tabular}{lccccccc}
        \toprule
         & \textbf{Training (h)} & \textbf{Batch size} & \multicolumn{3}{c}{\textbf{Inference (ms/sample)}} &  \textbf{\# Iterations/sample} \\
        Mesh res. (\# nodes)& & & \textbf{1069} & \textbf{5485} & \textbf{37362} & & \\
        \midrule
        \textbf{iFOL} & \textbf{14} & 64 &\textbf{1.26} & \textbf{1.60} & \textbf{3.59} & \textbf{3} \\
        \textbf{FEM} & - & 64 & {256} & {1599} & {29961} & 10 \\
        \midrule
         \textbf{AD-iFOL} & - &  32 & \textbf{169} & \textbf{173} & \textbf{228} &\textbf{0}\\
        \textbf{Adjoint-FEM} & - & 32 & 175 & 233 & 3152 &1\\
        \bottomrule
    \end{tabular}
\end{table}

\textcolor{black}{The FEM results reported in Table 2 were obtained using our own GPU-compatible FEM implementation in JAX, which provides additional speed-up and ensures consistency with the iFOL setup. All codes are openly available through the Folax framework \cite{folax2025github}, allowing readers to reproduce and verify the results. }

\color{black}

%

\color{black}

\color{black}
\subsection{iFOL vs. Baseline Neural Operators}
\label{sec:iFOL_vs_FNO_deepONet}
As illustrated in Fig.~\ref{fig:compare}, the architecture of the proposed iFOL framework differs from existing operator learning methods, specifically FNO and DeepONet, despite sharing certain similarities. Since the baseline FNO can only be applied to rectangular and cubic domains, we focus on the nonlinear hyper-elastic problem in Section \ref{sec:mechanics}. While the geometry is relatively simple, the governing PDE is highly nonlinear and the solution contains sharp discontinuities. These characteristics pose significant challenges for learning the mapping between the elasticity and displacement fields.

We trained the baseline variants of DeepONet \citep{Lu2021, Wang2021} and FNO \citep{li2023physicsinformedneuraloperatorlearning, ESHAGHI2025117785} in both data-driven and fully physics-informed manners. While other variants of DeepONet and FNO, such as those presented in \cite{LU2022114778}, are highly relevant, they were not explored in this study and are left for future investigation. To ensure a fair comparison, all key parameters, including the number of training samples, batch size, and optimizer type, were kept consistent across models. In the physics-informed setting, the loss function was identical for all models, with a backpropagation driven solely by nodal finite element residuals. Method-specific hyperparameters, such as activation functions, number of Fourier layers, and modes, were carefully tuned to optimize performance while maintaining fairness in the comparison. 

\begin{figure}[H]
  \centering  \includegraphics[width=1.0\linewidth]{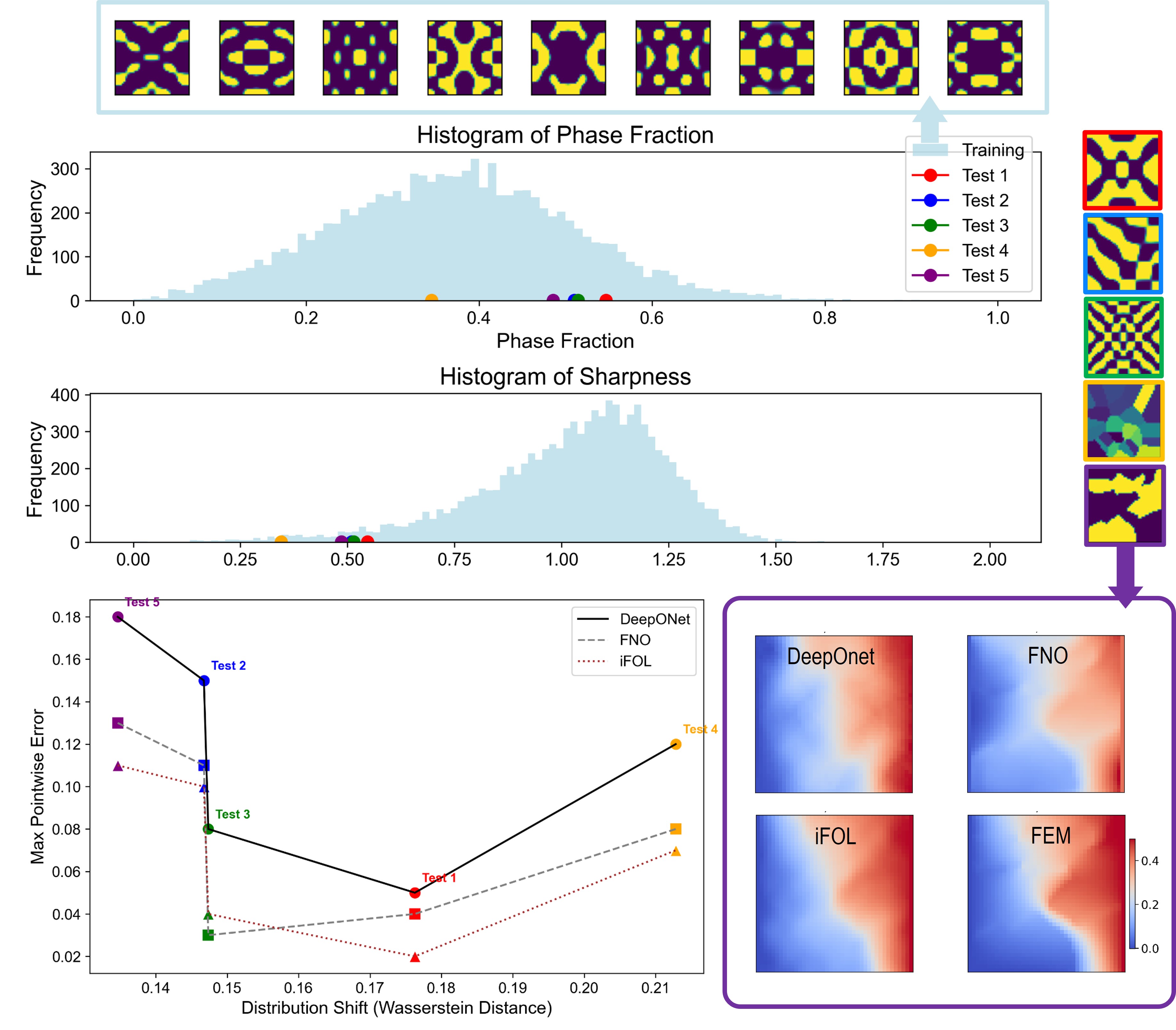}
  \caption{\textcolor{black}{A comparative analysis of operator learning architectures with a focus on their generalization performance. The iFOL framework is compared against FNO and DeepONet in the 2D nonlinear setting discussed in Section 4.1. } }
  \label{fig:compare_num}
\end{figure}

Figure~\ref{fig:compare_num} presents a comprehensive performance comparison of the neural operators under investigation. The upper panels present histograms of the training samples, accompanied by representative examples illustrating the corresponding phase fraction and sharpness values (see \cite{NajafiKoopas2025} for the mathematical equations and for further details of these quantities). For pointwise error analysis, five unseen test cases are selected and displayed on the right. Each case is inspired by a distinct application in materials engineering and represents a unique microstructural scenario. To enable a quantitative assessment of their topological relevance, the positions of these test samples are also indicated on the histograms. The bottom panel reports the average Wasserstein distance (\cite{peyre2019computational}) between each selected test sample and all 8,000 training samples, as well as the corresponding maximum pointwise error. The test case exhibiting the highest error is highlighted in the bottom-right plot.

Overall, the analysis demonstrates the strong performance of the iFOL framework compared to FNO and DeepONet. It is important to note that here we present the best-performing versions of each model: data-driven DeepONet, physics-informed FNO, and physics-informed iFOL. The physics-informed DeepONet did not yield promising results and consistently converged to trivial solutions for this challenging test problem.

In addition to the above analysis, a more comprehensive statistical evaluation was conducted using 100 unseen test samples drawn from two distinct categories. The first category consists of in-distribution samples that were not seen during training but share similar characteristics due to their symmetric and periodic Fourier-based parameterization. The second category includes out-of-distribution samples, generated using a dual-phase, unsymmetric Voronoi-based parameterization, thereby representing a more challenging generalization scenario.

The average errors for all models, evaluated separately for each test category, are summarized in Table~\ref{tab:network_err}. As observed consistently, the iFOL framework outperforms the other methods across both in-distribution and out-of-distribution test sets, followed by FNO and then DeepONet. Notably, iFOL achieves this superior performance with significantly fewer trainable model parameters compared to FNO.

\begin{table}[H]
    \centering
    \caption{\textcolor{black}{Error analysis of operator learning methods for the 2D nonlinear mechanical problem discussed in Sec.~\ref{sec:mechanics}.}}
    \label{tab:network_err}
    \begin{tabular}{llccc}
        \toprule
        \textbf{Method} & \textbf{Test Type} & \textbf{MAE} & \textbf{Max Point. Err.} & \textbf{\# Model Param.} \\
        \midrule
        \multirow{2}{*}{\textbf{iFOL}} 
            & IN~~~~~~~(sym. peri. Fourier-based) & \textbf{0.0061} & \textbf{0.0323} & \multirow{2}{*}{\textbf{143,938}} \\
            & OUT~~~~~(unsym. Voronoi-based) & \textbf{0.0355} & \textbf{0.1367} & \\
        \cmidrule{1-4}
        \multirow{2}{*}{\textbf{FNO}} 
            & IN & 0.0077 & 0.037 & \multirow{2}{*}{2,368,130} \\
            & OUT & 0.0465 & 0.1658 & \\
        \cmidrule{1-4}
        \multirow{2}{*}{\textbf{DeepONet}} 
            & IN & 0.0076 & 0.0533 & \multirow{2}{*}{149,506} \\
            & OUT & 0.0565 & 0.2316 & \\
        \bottomrule
    \end{tabular}
\end{table}

\color{black}
\section{Conclusion}
We introduced the implicit Finite Operator Learning (iFOL) technique, a novel framework for solving parametric PDEs on arbitrary geometries. iFOL is a physics-informed automatic encoder-decoder that combines three state-of-the-art techniques: conditional neural fields for spatial encoding, second-order meta-learning for PDE encoding, and a physics-informed loss function inspired by the original FOL concept. This operator learning technique is capable of providing both a precise continuous mapping between parameter and solution spaces and a reliable estimation of the operator's Jacobian. The computed Jacobian can be directly utilized for sensitivity analysis and gradient-based optimization, completely independent of the number of response functions. 

The methodology is validated across a range of PDEs, including hyperelasticity, linear elasticity, nonlinear thermal diffusion, and the Allen–Cahn equation, in both 2D and 3D settings with regular and complex geometries. These examples span stationary and transient problems in computational mechanics and materials science. In all cases, iFOL successfully learns parametric solutions and generalizes to unseen test samples without retraining. 

\textcolor{black}{It is worth highlighting that no labeled data is used throughout this work. Instead, the PDEs in the weighted residual form or in their energy counterparts are directly defined as loss functions.} This approach improves efficiency, as no automatic differentiation operator is required to construct the loss terms. Moreover, it allows for the direct integration of residuals from other numerical solvers, such as FEM or spectral methods like FFT. \textcolor{black}{Also it can be extended to other existing architectures, such as DeepONet and FNO. Our preliminary investigations indicate that iFOL benefits from FNO’s ability to capture high-frequency details, due to its SIREN-based architecture, while retaining the key advantage of DeepONet, its flexibility to handle complex geometries far beyond simplified cubic or square domains. Moreover, the number of trainable parameters is much less than the FNO version.} Finally, since the network learns the equation parametrically for a class of problems, the inference time of iFOL can be highly competitive with direct numerical solvers like FEM. The speed-up factor is not only problem-dependent but also resolution-dependent and influenced by implementation details, making it difficult to provide a universal comparison, especially considering differences in hardware setups for DL-based solvers and classical numerical methods. In specific cases and under comparable conditions, we observe at least a 100$\times$ speed-up for nonlinear problems, which can further increase for higher resolutions.

\textbf{Outlook and future research direction} 
A natural extension is to apply the introduced methodologies to more complex multiphysics problems, where multiple physical processes influence each other through one- or two-way coupling. In such scenarios, classical solvers often struggle and require staggered approaches.  
Moreover, in this work, we primarily rely on a single input parameter space at a time. A natural extension would be to incorporate multiple input parameter spaces, a direction that has been explored to some extent by other authors, primarily using data-driven operator learning techniques. 
\textcolor{black}{In this work, we employ ideas based on autoregressive neural networks for transient problems with a fixed time step size. This approach imposes certain limitations, particularly for rate-dependent problems, and does not offer temporal super-resolution capabilities, an aspect that should be addressed in future work.}
\textcolor{black}{Moreover, we acknowledge that abrupt changes in transient problems (e.g., fracture or turbulence) often benefit from adaptive time stepping. The proposed framework is compatible with established numerical strategies, including adaptive refinement. Since iFOL minimizes weighted residuals, the residual values can directly guide dynamic time-step refinement. The use of Adam for parametric residual minimization may offer stability advantages over Newton-type schemes in ill-conditioned problems. These aspects will be systematically investigated in future work.}

Finally, since we have access to meaningful and reliable sensitivity information independently of the number of objectives, gradient-based optimization becomes a compelling approach. This capability has significant potential for efficiently solving complex optimization and inverse problems, even in time-dependent settings, while maintaining reasonable computational and implementation costs.

\section*{Code Availability}
The implementation of iFOL and the examples presented in this study are publicly available under the Folax framework \cite{folax2025github}.

\section*{Acknowledgments}
The authors would like to thank the Deutsche Forschungsgemeinschaft
(DFG) for the funding support provided to develop the present work in the Cluster of Excellence Project 'Internet of Production' (project: 390621612). The authors also acknowledge the financial support of SFB 1120 B07 - Mehrskalige thermomechanische Simulation der fest-flüssig Interaktionen bei der Erstarrung (B07) (260070971) (260070971).
Yusuke Yamazaki acknowledges the financial support from JST SPRING (JPMJSP2123). Mayu Muramatsu acknowledges the financial support from the JSPS
KAKENHI Grant (22H01367).

\section*{Authors contributions}
R.N.A.: Conceptualization, Methodology, Software, Writing - Review \& Editing.
Y.Y.: Software, Writing - Review \& Editing. 
K.T.: Software.
M.M.: Funding, Resources, Review \& Editing. 
M.A.: Funding, Resources, Review \& Editing.
S.R.: Supervision, Methodology, Software, Writing - Review \& Editing.


\newpage
\appendix
\section{Review on physical formulations and their discretization using FEM}
\subsection{Mechanical problem}
\label{sec:mechanics_app}
Here, we summarize the mechanical problem where the position of material points is denoted by $\bm{X}^T=[x,~ y,~z]$. We denote the displacement components by $U$, $V$, and $W$ in the $x$, $y$, and $z$ directions, respectively. 
The kinematic relation defines the strain tensor $\bm{\varepsilon}$ in terms of the deformation vector $\bm{U}^T=[U,~ V,~W]$ and reads:
\begin{align}
\label{kinematics}
\bm{\varepsilon}\,&=\,\text{sym}(\text{grad}(\bm{U})) = \nabla^s\bm{U} = \dfrac{1}{2}\left(\nabla \bm{U} + \nabla \bm{U}^T \right),\\
\boldsymbol{F} &= \dfrac{\partial \bm{x}}{\partial \bm{X}},~\boldsymbol{C} = \boldsymbol{F}^T \boldsymbol{F},~J_F = \det(\boldsymbol{F}).
\end{align} 
Here, \( \boldsymbol{F} \) is the deformation gradient, and \( J_F \) is the Jacobian determinant, representing the local volume change. 
The elastic energy of the solid for linear and nonlinear (hyperelastic Neo-Hookean material) case reads as
\begin{align}
\label{lin_energy}  
W_{lin} &= \frac{\lambda}{2} (\text{tr} \boldsymbol{\varepsilon})^2 + \mu \, \text{tr} (\boldsymbol{\varepsilon}^2), \\
W_{nonlin} &= \frac{\mu(\bm{X})}{2} (\bar{I_1} - 3) - \frac{\kappa(\bm{X})}{4}(J_F^2 - 1 - 2 \ln J_F).
\end{align}
The Lamé constants are denoted by \( \lambda \) and \( \mu \), and \( \text{tr}(\boldsymbol{\varepsilon}) \) is the trace of the strain tensor, representing the volumetric strain.
\( \bar{I}_1 = \text{tr}(\mathbf{\bar{C}}) \) is the first invariant of the isochoric right Cauchy-Green deformation tensor \( \mathbf{\bar{C}} = \mathbf{J}^{-2/3} \mathbf{C}\) where \( \mathbf{C} \) is the right Cauchy-Green tensor. 
We then define the Cauchy stress tensor $\bm{\sigma} = \frac{\partial W_{lin}}{\partial \mathbf{\epsilon}} = {\mathbb{C}}~{\bm{\varepsilon}}$ and the Piola-Kirchhoff stress $\boldsymbol{P} = \frac{\partial W_{nonlin}}{\partial \mathbf{F}}$, which are obtained by differentiating the strain energy function. 
Here $\mathbb{C}(\lambda, \mu)= \lambda~\mathbf{I} \otimes \mathbf{I} + 2\mu~\mathbb{I}^s$ or $\mathbb{C}(E, \nu)$ is the fourth-order elasticity tensor by defining $\mathbf{I}$ as the second-order identity tensor and $\mathbb{I}^s$ as the symmetric fourth-order identity tensor. 
Finally, the mechanical equilibrium in the absence of body force, as well as the Dirichlet and Neumann boundary conditions, are written as:
\begin{align}
\label{Equilbrium}
\text{div}({\bm{\sigma}}) = \text{div}(\mathbb{C}\nabla^s\bm{U})\ &= \bm{0}~~~~\text{in}~~~ \Omega\\
\label{BcsMech_d}
\bm{U} &= \bar{\bm{U}}~~~\text{on}~~\Gamma_D \\ 
\label{BcsMech_n}
\bm{\sigma} \cdot \bm{n} = \bm{t} &= \bar{\bm{t}}~~~~\text{on}~~\Gamma_N
\end{align} 
In the above relations, $\Omega$ and $\Gamma$ denote the material points in the body and on the boundary area, respectively. Moreover, the Dirichlet and Neumann boundary conditions are introduced in Eq.~\ref{BcsMech_d} and Eq.~\ref{BcsMech_n}, respectively.
The variables \( \hat{\boldsymbol{\sigma}} \) and \( \boldsymbol{C} \) in Table \ref{tab:sum} represent the stress and stiffness tensors in Voigt notation, respectively.
In the case of hyperelasticity, we then have 
$\text{Div}({\bm{P}}) = \bm{0}$ instead of Eq.~\ref{Equilbrium}.

\subsection{Thermal problem}
\label{sec:thermal_app}
The heat equation describes how the temperature \( T(\bm{x}, t) \), with \( \bm{x} \) representing position and \( t \) the time, evolves within the domain \( \bm{x} \in \Omega \) over time.  
Let the heat source be \( Q: \Omega \times (0, \tau) \to \mathbb{R} \),  
the boundary temperature \( T_D(\bm{x}): \Gamma_D \times (0, \tau) \to \mathbb{R} \),  
and the boundary heat flux be \( q_N: \Gamma_N \times (0, \tau) \to \mathbb{R} \),  
where \( \Gamma_D \) and \( \Gamma_N \) are the regions where the Dirichlet and Neumann boundary conditions are applied, respectively. Here, \( t \in (0, \tau) \) represents the temporal domain, with \( \tau \) denoting the end time. The strong form of the heat equation is given as:
\begin{align}
c \rho  \frac{\partial T(\bm{x}, t)}{\partial t} &= -\text{div}(\bm{q}) + Q \quad \text{in} \ \Omega \times (0, \tau), \label{eq:tran_therm} \\
T(\bm{x}, t) &= T_D(\bm{x})  \quad \text{on} \ \Gamma_D \times (0, \tau), \\
\nabla T(\bm{x}, t) \cdot \bm{n} &= q_N(\bm{x})  \quad \text{on} \ \Gamma_N \times (0, \tau), \\
T(\bm{x}, 0) &= T_0(\bm{x}) \quad \text{in} \ \Omega.
\end{align}

In the above equations, \( \bm{n} \) is the outward normal vector, \( c \) is the specific heat capacity, and \( \rho \) is the density. The heat flux is given by \( \boldsymbol{q} = -K \nabla T \), where \( K = k_0(\bm{x})(1+\alpha T) \) represents the thermal conductivity, which depends on both position and temperature.  

The above set of equations forms the basis for the results and studies presented in Section \ref{sec:thermal}.  
For the stationary thermal problem (see Section \ref{sec:FOL_geom}), the time derivative in Eq.~\ref{eq:tran_therm} naturally vanishes.

\subsection{Allen-Cahn equation}
\label{sec:allen_cahn_app}
The Allen-Cahn equation describes the evolution of an order parameter \( \phi(\bm{x}, t) \), where \( \bm{x} \) represents position and \( t \) is time, within the domain \( \bm{x} \in \Omega \).  
Let the source term be \( S: \Omega \times (0, \tau) \to \mathbb{R} \),  
the boundary value be \( \phi_D(\bm{x}): \Gamma_D \times (0, \tau) \to \mathbb{R} \),  
and the boundary flux be \( q_N: \Gamma_N \times (0, \tau) \to \mathbb{R} \),  
where \( \Gamma_D \) and \( \Gamma_N \) are the regions where the Dirichlet and Neumann boundary conditions are applied, respectively. Here, \( t \in (0, \tau) \) represents the temporal domain, with \( \tau \) denoting the end time. The strong form of the Allen-Cahn equation is given as:
\begin{align}
\frac{\partial \phi(\bm{x}, t)}{\partial t} &= - M \left( \frac{\delta F}{\delta \phi} \right)  \quad\quad\quad\quad\quad\quad \text{in} \ \Omega \times (0, \tau), \label{eq:allen_cahn} \\
F &= \int_{\Omega} \left( \frac{\epsilon}{2} |\nabla \phi|^2 + f(\phi) \right) d\Omega~~~~\text{in}~~~ \Omega \times (0, \tau), \\
\phi(\bm{x}, t) &= \phi_D(\bm{x})  ~~\quad\quad\quad\quad\quad\quad\quad\quad \text{on} \ \Gamma_D \times (0, \tau), \\
\nabla \phi(\bm{x}, t) \cdot \bm{n} &= \phi_N(\bm{x})  ~~\quad\quad\quad\quad\quad\quad\quad\quad \text{on} \ \Gamma_N \times (0, \tau), \\
\phi(\bm{x}, 0) &= \phi_0(\bm{x}) ~~~\quad\quad\quad\quad\quad\quad\quad\quad \text{in} \ \Omega.
\end{align}

In the above equations, \( \bm{n} \) is the outward normal vector, \( M \) is the mobility coefficient, and \( \frac{\delta F}{\delta \phi} \) represents the functional derivative of the free energy, typically given by:
\begin{align}
    \frac{\delta F}{\delta \phi} = -\epsilon^2 \nabla^2 \phi + f^\prime(\phi),
 \end{align}
where \( \epsilon \) is a small parameter controlling the interface thickness and  $f^\prime(\phi)= (\phi^2-1)\phi$ denotes the derivative of the double-well potential enforcing phase separation.  
The above set of equations forms the basis for the results and studies presented in Section \ref{sec:allen_cahn}.  
The Allen-Cahn equation is a fundamental phase-field model and plays a crucial role in modeling microstructure evolution in alloys \cite{ALLEN19791085, Karma1998}. The equation captures the dynamics of interfaces and transitions between phases (including applications in phase-field fracture \cite{BOURDIN2000797, REZAEI2021104253}). Its applications extend to image processing, tumor growth modeling, and topology optimization.

\subsection{A short review on FEM formulation}
\label{sec:review_FEM}
Next, we briefly summarize the FEM discretization techniques, particularly for a 2D quadrilateral element, to clarify the loss terms in the iFOL formulation, especially for implementation purposes. The details provided here are fairly standard, and the extension to other types of complex element formulations with different polynomial orders should be straightforward. Readers are also encouraged to see the standard procedure in any finite element subroutine \cite{hughes2000finite, bathe1996finite}. The corresponding linear shape functions $\bm{N}$ and the deformation matrix $\bm{B}$ used to discretize the mechanical weak form in the current work.
\begin{align}
\label{eq:N}
\bm{N} =
\begin{bmatrix}
N_1 & 0 & \dots & N_4 & 0\\
0 & N_1 & \dots & 0 & N_4
\end{bmatrix},~
\bm{B} =
\begin{bmatrix}
N_{1,x} & 0       & \dots & N_{4,x} & 0        \\
0       & N_{1,y} & \dots & 0       & N_{4,y}  \\
N_{1,y} & N_{1,x} & \dots & N_{4,y} & N_{4,x} 
\end{bmatrix}.
\end{align}
The notation $N_{i,x}$ and $N_{i,y}$ represent the derivatives of the shape function $N_i$ with respect to the coordinates $x$ and $y$, respectively. To compute these derivatives, we utilize the Jacobian matrix
\begin{align}
 \boldsymbol J = \partial \boldsymbol X / \partial \boldsymbol \xi = \begin{bmatrix} \frac{\partial x}{\partial \xi} & \frac{\partial y}{\partial \xi} \\ \frac{\partial x}{\partial \eta} & \frac{\partial y}{\partial \eta} \end{bmatrix}. 
\end{align}

Here, $\boldsymbol X = [x, y]$ and $\boldsymbol \xi = [\xi, \eta]$ represent the physical and parent coordinate systems, respectively. It is worth mentioning that this determinant remains constant for parallelogram-shaped elements, eliminating the need to evaluate this term at each integration point.
Finally, for the B matrix, we have 
\begin{align}
 \bm{B} = \bm{J}^{-1} \begin{bmatrix} \frac{\partial N_1}{\partial \xi} & 0 & \frac{\partial N_2}{\partial \xi} & 0 & \frac{\partial N_3}{\partial \xi} & 0 & \frac{\partial N_4}{\partial \xi} & 0 \\ 0 & \frac{\partial N_1}{\partial \eta} & 0 & \frac{\partial N_2}{\partial \eta} & 0 & \frac{\partial N_3}{\partial \eta} & 0 & \frac{\partial N_4}{\partial \eta} \\ \frac{\partial N_1}{\partial \eta} & \frac{\partial N_1}{\partial \xi} & \frac{\partial N_2}{\partial \eta} & \frac{\partial N_2}{\partial \xi} & \frac{\partial N_3}{\partial \eta} & \frac{\partial N_3}{\partial \xi} & \frac{\partial N_4}{\partial \eta} & \frac{\partial N_4}{\partial \xi} \end{bmatrix}. 
\end{align}

For each element the deformation field and stress tensor $\hat{\bm{\sigma}}$ are approximated as
\begin{align}
U = \sum {N}_i u_i =\boldsymbol{N} \boldsymbol U_e,~
V = \sum {N}_i V_i =\boldsymbol{N} \boldsymbol V_e,~ 
\hat{\bm{\sigma}} &= \bm{C}\boldsymbol{B} \bm{U}_e, 
\end{align}
Here, $\boldsymbol U^T_e=\left[
\begin{matrix}
u_1  &\cdots & u_4 \\
\end{matrix}
\right]$ and $\bm{V}_e$ are the nodal values of the deformation field of element $e$ in the $x$ and $y$ directions, respectively. 
The same procedure applies to both the thermal problem and the Allen-Cahn equation. However, one must account for additional time derivatives. We summarize the implemented loss terms in Table \ref{tab:fe_loss}.


\begin{table}[H]
\caption{Summary of the FEM-based residuals of the investigated problems.}  
\centering
{ 
    \renewcommand{\arraystretch}{2.0} 
\begin{tabular}{ l l l }
\hline
\hline
PDE type  & FEM residual / iFOL loss function \\
\hline
\makecell[l]{Stationary \\ nonlinear mechanical \\ \\ 
$\boldsymbol{C}(x,y) \to \boldsymbol{U}(x,y)$ \\
$\mathbf{u}^e_{\theta, \gamma} \equiv \boldsymbol{U}_e$, (Sec.~\ref{sec:mechanics}) }  
& \makecell[l]{
$r_{e} = -\int_{\Omega_e} \boldsymbol{S} \colon \delta \boldsymbol{E} dV - \int_{\Omega_e}(\Delta \boldsymbol{S} \colon \delta \boldsymbol{E} + \boldsymbol{S} \colon \Delta \delta \boldsymbol{E}) dV + \int_{\Omega_t}\delta \boldsymbol{U}.~ \boldsymbol{T} dA$} \\
&  \makecell[l]{$\mathcal{L}_{iFOL}= \sum_{e=1}^{n_{el}} \sum_{k=1}^{n_{int}}~\left(\frac{\mu(\boldsymbol \xi_k)}{2}~\left(\text{tr}(\bar{\boldsymbol{C}}(\boldsymbol{U_e})) - 3\right)~+~\frac{\kappa(\boldsymbol \xi_k)}{4}~(J_F^2(\boldsymbol{U_e}) - 2\ln J_F(\boldsymbol{U_e}) - 1)\right)~\text{det}(\boldsymbol{J})~w_k$} \\ 
\hdashline
\makecell[l]{Stationary \\ linear mechanical \\ \\ 
$\boldsymbol{U}_b \to \boldsymbol{U}(x,y)$ \\
$\mathbf{u}^e_{\theta, \gamma} \equiv \boldsymbol{U}_e$, (Sec.~\ref{sec:mechanics_bcs}) }
&
\makecell[l]{$r_{e}= -\int_{\Omega_e} [\boldsymbol B]^T \boldsymbol C~[\boldsymbol B]\boldsymbol U_e ~dV + \int_{\Omega_t}[\boldsymbol N]^T \boldsymbol C~[\boldsymbol B \boldsymbol U_e]^T ~\boldsymbol n~dS$} \\
&  
\makecell[l]{$\mathcal{L}_{iFOL}= \sum_{e=1}^{n_{el}} \boldsymbol U^T_e \left[ \boldsymbol \left(\sum_{k=1}^{n_{int}} \dfrac{w_k}{2}~\text{det}(\boldsymbol J)~[\boldsymbol B(\boldsymbol \xi_k)]^T \boldsymbol C_e(\boldsymbol \xi_k) \boldsymbol B(\boldsymbol \xi_k) \right)~\boldsymbol U_e \right].$} \\
\hdashline
\makecell[l]{Stationary \\ nonlinear thermal \\ \\
$K(x,y) \to T(x,y)$ \\
$\mathbf{u}^e_{\theta, \gamma} \equiv \boldsymbol T_e$, (Sec.~\ref{sec:FOL_geom}) }  &  \makecell[l]{$r_{e}= - \int_{\Omega_e} [\boldsymbol B]^T \boldsymbol K(\boldsymbol T_e)~[\boldsymbol B]\boldsymbol T_e ~dV + \int_{\Omega_t}[\boldsymbol N]^T \boldsymbol t ~dS + \int_{\Omega_e}[\boldsymbol N]^T Q ~dV$ } \\ 
&  \makecell[l]{$\mathcal{L}_{iFOL}= \sum_{e=1}^{n_{el}} \left[\boldsymbol T_e^{n+1}\right]^T \left[ \boldsymbol \left(\sum_{k=1}^{n_{int}} \dfrac{w_k}{2}~\text{det}(\boldsymbol J)~[\boldsymbol B(\boldsymbol \xi_k)]^T  \boldsymbol K(\boldsymbol T_e^{n+1},\boldsymbol \xi_k) \boldsymbol B(\boldsymbol \xi_k) \right)~\boldsymbol T_e^{n+1}  \right]$} \\
\hdashline
\makecell[l]{Transient \\ nonlinear thermal \\ \\ 
$T^{n}(x,y) \to T^{n+1}(x,y)$ \\
$\mathbf{u}^e_{\theta, \gamma} \equiv \boldsymbol T_e^{n+1}$, (Sec.~\ref{sec:thermal})}   &  \makecell[l]{$r_{e}= - \int_{\Omega_e} [\boldsymbol B]^T \boldsymbol K(\boldsymbol T_e^{n+1})~[\boldsymbol B]\boldsymbol T_e^{n+1} ~dV + \int_{\Omega_t}[\boldsymbol N]^T \boldsymbol t ~dS + \int_{\Omega_e}[\boldsymbol N]^T Q ~dV$ \\ 
~~~~~~~~~$- \int_{\Omega_e} [\boldsymbol N]^T \rho c_p ~[\boldsymbol N] \frac{\boldsymbol T_e^{n+1}- \boldsymbol T_e^{n}}{\Delta t } ~dV$}  \\
&  \makecell[l]{$\mathcal{L}_{iFOL}= \sum_{e=1}^{n_{el}} \left[\boldsymbol T_e^{n+1}\right]^T \left[ \boldsymbol \left(\sum_{k=1}^{n_{int}} \dfrac{w_k}{2}~\text{det}(\boldsymbol J)~[\boldsymbol B(\boldsymbol \xi_k)]^T  \boldsymbol K(\boldsymbol T_e^{n+1},\boldsymbol \xi_k) \boldsymbol B(\boldsymbol \xi_k) \right)~\boldsymbol T_e^{n+1}  \right]$\\ ~~~~~~~~~~~~~ + $\sum_{e=1}^{n_{el}}  \left[  \sum_{k=1}^{n_{int}} \dfrac{w_k}{2}~\text{det}(\boldsymbol J)~\rho c_p \left[\boldsymbol N(\boldsymbol \xi_k)\right]^T\frac{\left(\boldsymbol N(\boldsymbol \xi_k)\boldsymbol T_e^{n+1}-\boldsymbol N(\boldsymbol \xi_k)\boldsymbol T_e^{n}\right)^2 }{\Delta t
}  \right]. $}  \\
\hdashline
\makecell[l]{ Transient \\ Allen-Cahn \\ \\ 
$\phi^{n}(x,y) \to \phi^{n+1}(x,y)$ \\
$\mathbf{u}^e_{\theta, \gamma} \equiv \boldsymbol \phi_e^{n+1}$, (Sec.~\ref{sec:allen_cahn}) }   &  \makecell[l]{$r_{e}= - \int_{\Omega_e} [\boldsymbol B]^T \boldsymbol [\boldsymbol B]\boldsymbol \phi_e^{n+1} ~dV - \int_{\Omega_e} [\boldsymbol N]^T ~[\boldsymbol N] \frac{\boldsymbol \phi_e^{n+1}- \boldsymbol \phi_e^{n}}{\Delta t } ~dV$ \\
~~~~~~~~~$- \int_{\Omega_e}[\boldsymbol N]^T\frac{1}{\varepsilon^2} \left[\left(\boldsymbol{N}\boldsymbol{\phi}_e^{n+1}\right)^2-1\right]\boldsymbol N\boldsymbol{\phi}_e^{n+1}~dV + \int_{\Omega_t}[\boldsymbol N]^T \left( \nabla \boldsymbol{\phi}_{e}^{n+1}\cdot \boldsymbol{n} \right)~dS.$}  \\
&  \makecell[l]{$\mathcal{L}_{iFOL}= 
\sum_{e=1}^{n_{el}} \left[\boldsymbol \phi_e^{n+1}\right]^T \left[ \boldsymbol \left(\sum_{k=1}^{n_{int}} \dfrac{w_k}{2}~\text{det}(\boldsymbol J)~[\boldsymbol B(\boldsymbol \xi_k)]^T  \boldsymbol B(\boldsymbol \xi_k) \right)~\boldsymbol \phi_e^{n+1} \right]$\\ 
~~~~~~~~~~~~~$ + \sum_{e=1}^{n_{el}}  \left[  \sum_{k=1}^{n_{int}} \dfrac{w_k}{2}~\text{det}(\boldsymbol J)~[\boldsymbol N(\boldsymbol \xi_k)]^T \frac{\left( \boldsymbol N(\boldsymbol \xi_k)\boldsymbol \phi_e^{n+1}-\boldsymbol N(\boldsymbol \xi_k)\boldsymbol \phi_e^{n} \right)^2}{\Delta t
}   \right]$ \\ 
~~~~~~~~~~~~~$+ \sum_{e=1}^{n_{el}}  \left[ \boldsymbol \sum_{k=1}^{n_{int}} w_k~\text{det}(\boldsymbol J)~\frac{1}{\varepsilon^2} \dfrac{\left[( \boldsymbol N(\boldsymbol \xi_k) \boldsymbol{\phi}_e^{n+1})^2-1\right]^2}{4}\right].
$
}\\
\hline
\end{tabular}
}

\label{tab:fe_loss}
\end{table} 


In all the reported studies, we utilized simple first-order shape functions. The element type used in the studies varies based on the application, ranging from structured to unstructured meshes, including quadrilateral and tetrahedral elements, demonstrating the flexibility of the approach in handling problems with complex geometries.

Moreover, in the above set of equations, $n_{\text{int}}$ and $n_{\text{el}}$ represent the number of Gaussian integration points and number of elements, respectively. $\boldsymbol{\xi}_k$ and $w_k$ denote the coordinates and weight of the $k$-th integration point. The determinant of the Jacobian matrix is denoted by $\text{det}(\boldsymbol{J})$. Readers are encouraged to refer to \cite{bathe1996finite,hughes2000finite} as well as \cite{rezaei2025finite, yamazaki2025finite} and the references therein for more details on the FE formulation.

\newpage
\section{Choice of Hyperparameters}
\label{sec:implementation_details}

A comprehensive summary of the network's hyperparameters and configurations for all the models introduced above, as well as the various analyses conducted, is provided in Tables~\ref{tab:hyperparam_stan} and \ref{tab:hyperparam_tran}. NVIDIA Quadro RTX 6000s with 24 GB of RAM and NVIDIA A100 with 40 GB of RAM were utilized for the training of the stationary problems. The studies on the transient problems were performed on NVIDIA GeForce RTX 4090 with 24 GB of RAM.

We conducted extensive studies on multiple parameters, including latent size, learning rate, neural network structure, and the frequency of sinusoidal functions, among others. The values reported in Tables \ref{tab:hyperparam_stan} and \ref{tab:hyperparam_tran} correspond to the best-performing configurations for iFOL. In general, we did not observe significant improvements with frequency parameter \(\omega_0\) other than $30$, which interestingly performed well across various problem classes. Increasing the depth and latent size proved consistently beneficial with the sinusoidal activation function used in this study, though it comes with a higher training cost. Depending on problem complexity, reducing the number of epochs and latent iterations can further accelerate both training and inference in the proposed iFOL method. As discussed in the results section, having a larger number of samples also leads to better performance. On the other hand, the training cost may increase depending on the chosen batch size, and one must adapt the network architecture and latent size accordingly.

\begin{table}[H]
\centering
 \caption{Network hyperparameters for stationary problems}
 \begin{tabular}{l l l l l}
    \hline
         Training parameter &  Nonlin. elas. (Sec.~\ref{sec:mechanics})
 & Lin. elas. (Sec.~\ref{sec:mechanics_bcs}) & Staion. Nonlin. therm. (Sec.~\ref{sec:FOL_geom})  \\
    \hline
     Number of samples & 8000 & 1000 &8000\\    
     Grid in training & $41 \times 41$ & $5605$ & $5485$\\
     Grid in evaluation & $101 \times 101$ & $5605$ & 37362\\
     Synthesizers  & [64]*4 & [256]*6 & [128]*6\\
     \(\omega_0\) & 30 & 30 & 30\\
     Modulators & Linear (FiLM) & Linear (FiLM) & Linear (FiLM)\\
     Latent size & 512 & 1024 & 512\\
     Number of latent iterations & 3 & 3 & 3\\     
     Latent/encoding learning rate & $10^{-2}$ & $10^{-2}$ & $10^{-2}$\\     
     Training learning rate & $10^{-5}$ & $10^{-4}$ to $ 10^{-7}$ & $10^{-5}$\\
     Batch size & 320 & 120 & 100\\
     Gradient normalization & Yes & Yes & Yes\\
     Number of epochs & 10000 & 10000 & 3800\\
     Optimizer& Adam & Adam & Adam\\     
    \hline
    Total trainable parameters & 143938 & 330755 & 476417\\
    \hline\\
    \end{tabular}
    \label{tab:hyperparam_stan}
\end{table}

\begin{table}[H]
\centering
 \caption{Network hyperparameters for the transient problems}
 \begin{tabular}{l l l}
    \hline
         Training parameter  &  Transient nonlin. thermal (Sec.~\ref{sec:thermal}) & Transient nonlin. Allen-Cahn (Sec.~\ref{sec:allen_cahn}) \\
    \hline
     Number of samples & 6000 & 8000 \\
     Grid in training & $51 \times 51$ & 2624\\
     Grid in evaluation & $51 \times 51$, $101 \times 101$ & 2624, 5678, 9873 \\     
     Synthesizers & [256]*6 & [256]*6\\
     \(\omega_0\) & 30 & 30\\
     Modulators & Linear (FiLM) & Linear (FiLM)\\
     Latent size & 256 & 256\\
     Number of latent iterations & 3 & 3 \\
     Latent/encoding learning rate & $1.0 \times 10^{-2}$ & $1.0 \times 10^{-2}$ \\
     Training learning rate & $1.0 \times 10^{-4} $ to $1.0 \times 10^{-7}$ & $1.0 \times 10^{-4} $ to $1.0 \times 10^{-7}$\\
     Batch size & 120 & 100\\
     Gradient normalization & Yes & Yes \\
     Number of epochs & 10000 & 10000 \\
     Optimizer & Adam & Adam \\
    \hline
    Total trainable parameters & 723457 & 723201\\
    \hline\\
    \end{tabular}
    \label{tab:hyperparam_tran}
\end{table}

\color{black}
Furthermore, for the baseline FNO and DeepONet neural operators studied in Section \ref{sec:iFOL_vs_FNO_deepONet}, we performed hyperparameter tuning and method optimization. The vanilla DeepONet network for data-driven training was found to perform best with both the trunk and branch networks implemented as MLPs consisting of four hidden layers with 64 neurons each, followed by an output layer with 128 neurons and ReLU activation functions. Since the network predicts a two-dimensional vector-valued displacement field, we adopted the split strategy for both the trunk and branch networks, as described in \cite{LU2022114778}. Specifically, each of the branch and trunk networks produces 128 output neurons. The dot product between the first 64 output neurons of the branch and trunk networks yields the \(x\) component of the displacement, while the dot product of the remaining 64 neurons yields the \(y\) component.

The FNO network used in this work is based on the Fourier Neural Operator architecture proposed by \cite{li2021fourierneuraloperatorparametric}. A linear transformation is first applied to lift the input to 32 channels, followed by four Fourier layers, each utilizing 12 Fourier modes in both the \(x\), \(y\) directions. Finally, a linear layer with 128 channels is used for back-projection. GELU activation functions are applied throughout the network.
\color{black}


\color{black}
\section{Insight on DL-based solvers}
\label{sec:DL_solver}
As emphasized throughout the manuscript, one of the key novelties of the proposed approach is the integration of classical numerical methods—such as the Finite Element Method (FEM)—into the neural network training process via a physics-informed loss function. In particular, the formulation embeds the enforcement of boundary conditions directly into the loss, mirroring how they are handled in classical solvers.


To highlight this aspect, we consider a simplified setting where the network is trained on a single parametric instance, minimizing the corresponding residuals derived from the governing physics. In this special case, the framework reduces to a Physics-Informed Neural Network, but instead of relying on automatic differentiation, it minimizes algebraic residuals obtained from standard numerical discretizations. As a result, one can expect the network to converge more precisely to the classical numerical solution (e.g., FEM) for the given instance.

\begin{figure}[H]
  \centering  \includegraphics[width=1.0\linewidth]{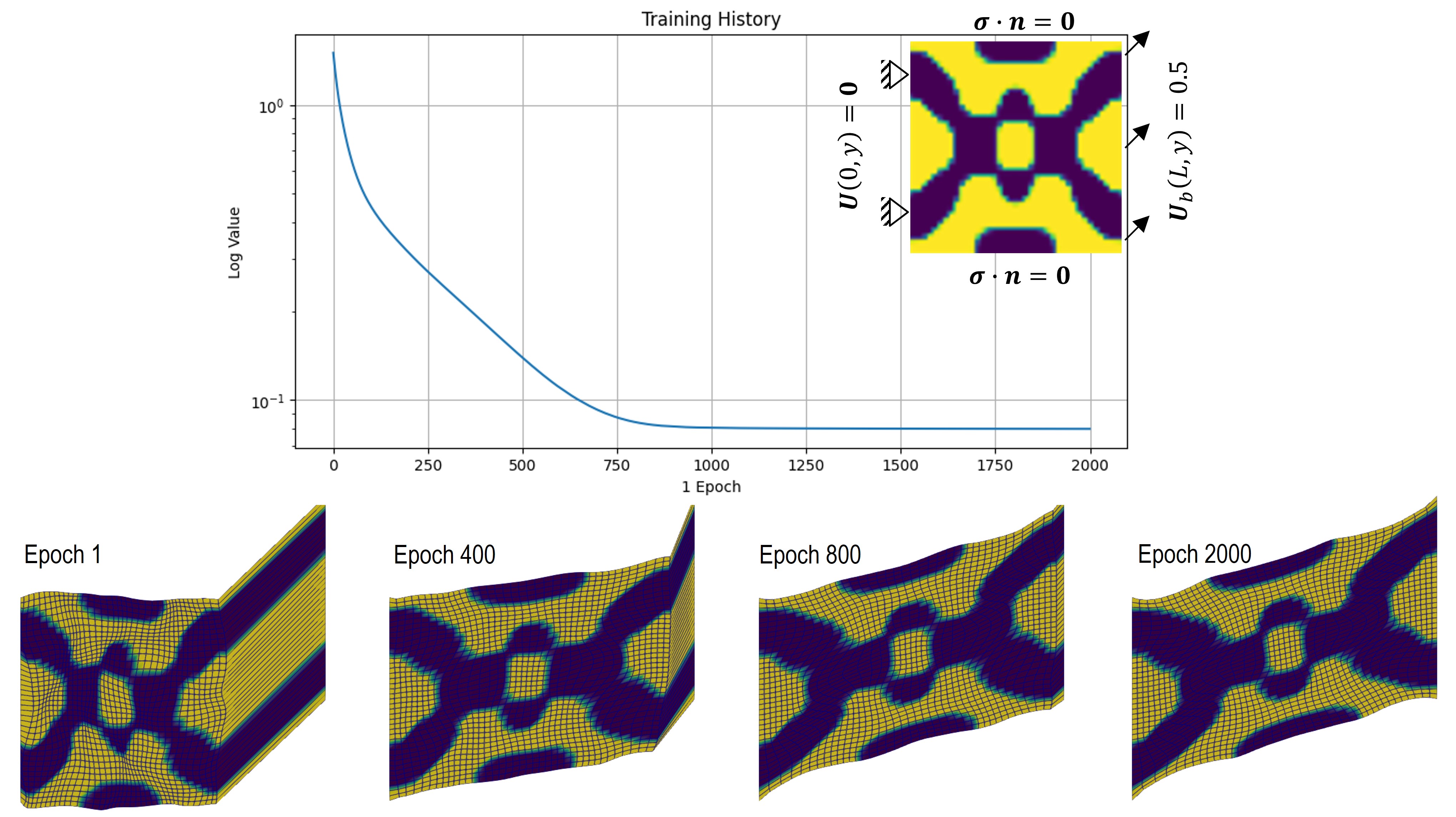}
  \caption{\textcolor{black}{ Evolution of the solution field for the nonlinear mechanical problem described in section \ref{sec:mechanics}. } }
  \label{fig:OTF}
\end{figure}

This capability is demonstrated in Fig.\ref{fig:OTF}, where we revisit the test case discussed in Section\ref{sec:mechanics} using the same model setup and boundary conditions. The evolution of the loss during training and the predicted solution snapshots are presented. Notably, the Dirichlet boundary conditions are strictly satisfied from the beginning of training, as they are explicitly encoded in the formulation. The remaining optimization focuses solely on learning the balance of internal and external forces, demonstrating the strength of this physics-aligned learning strategy.

\color{black}

\newpage
\bibliographystyle{unsrtnat}
\bibliography{references}  

\end{document}